\documentclass[fleqn,10pt]{wlscirep}
\usepackage[utf8]{inputenc}
\usepackage[T1]{fontenc}
\usepackage{subcaption} 
\usepackage{algorithm}%
\usepackage{algorithmicx}%
\usepackage{algpseudocode}%

\title{Enhancing Epidemic Forecasting: Evaluating the Role of Mobility Data and Graph Convolutional Networks}

\author[1,2]{Suhan Guo}
\author[1,2]{Zhenghao Xu}
\author[1,2*]{Furao Shen}
\author[1,3*]{Jian Zhao}
\affil[1]{Nanjing University, State Key Laboratory for Novel Software Technology, Nanjing, 210023, China}
\affil[2]{Nanjing University, School of Artificial Intelligence, Nanjing, 210023, China}
\affil[3]{Nanjing University, School of Electronic Science and Engineering, Nanjing, 210023, China}
\affil[*]{frshen@nju.edu.cn}


\begin{abstract}
Accurate prediction of contagious disease outbreaks is vital for informed decision-making. Our study addresses the gap between machine learning algorithms and their epidemiological applications, noting that methods optimal for benchmark datasets often underperform with real-world data due to difficulties in incorporating mobility information. We adopt a two-phase approach: first, assessing the significance of mobility data through a pilot study, then evaluating the impact of Graph Convolutional Networks (GCNs) on a transformer backbone. Our findings reveal that while mobility data and GCN modules do not significantly enhance forecasting performance, the inclusion of mortality and hospitalization data markedly improves model accuracy. Additionally, a comparative analysis between GCN-derived spatial maps and lockdown orders suggests a notable correlation, highlighting the potential of spatial maps as sensitive indicators for mobility. Our research offers a novel perspective on mobility representation in predictive modeling for contagious diseases, empowering decision-makers to better prepare for future outbreaks.
\end{abstract}
\begin{document}

\flushbottom
\maketitle
%
%
\thispagestyle{empty}

\section*{Introduction}
Accurate forecasting of daily case numbers for infectious diseases is crucial for disease management. The exponential growth of cases serves as an indicator of upcoming waves and the emergence of new variants \cite{elliottExponentialGrowthHigh2021}. Such forecasts allow governments to update and implement directives, facilitating proactive resource allocation to prevent strain on healthcare systems.

Airborne infectious disease outbreaks heavily rely on frequent human interaction that large-scale social activities significantly amplifying the transmission of communicable diseases \cite{colizzaRoleAirlineTransportation2006, halloranEbolaMobilityData2014, brockmannHiddenGeometryComplex2013, balcanMultiscaleMobilityNetworks2009, jiaPopulationFlowDrives2020}. The global impact of SARS-CoV-2 is unparalleled compared to previous coronavirus pandemics \cite{abdelrahmanComparativeReviewSARSCoV22020}, owing to its highly contagious nature and the increased accessibility of both intra- and inter-national travel facilitated by enhanced global connectivity \cite{findlaterHumanMobilityGlobal2018}. International travels across continents promote the dissemination of pathogens, ultimately resulting in pandemics \cite{wilsonTravellersGiveWings2020, chenStatespecificProjectionCOVID192020}. Extensive research has demonstrated that human mobility plays a pivotal role in driving the geographic spread of infectious diseases. Consequently, the routine movement of individuals between sub-regions is intertwined with their epidemic phases, necessitating a spatial perspective in disease forecasting \cite{ruktanonchaiAssessingImpactCoordinated2020, meloniModelingHumanMobility2011, sillsAggregatedMobilityData2020, kraemerEffectHumanMobility2020}. 

The use of mobility data as a proxy for effective contact has raised concerns regarding its efficacy in capturing its correlation with increased disease transmission \cite{gataloAssociationsPhoneMobility2021}. Nevertheless, mobility data remains a crucial source for modeling disease transmission in government surveillance \cite{birrellRealtimeNowcastingForecasting, ScientificEvidenceSupporting} and epidemic research \cite{wardForecastingSARSCoV2Transmission2022, nouvelletReductionMobilityCOVID192021, badrAssociationMobilityPatterns2020, liuAssociationsChangesPopulation2021}. Real-time behavioral data, such as surveys on social media platforms and search term volumes, are employed to describe mobility patterns, yet their collection poses challenges due to the inherent difficulty in tracking human behavior \cite{nixonRealtimeCOVID19Forecasting2022}. The resulting datasets often exhibit limited geographical coverage and contain numerous missing data points, thereby diminishing the effectiveness of mobility data as a proxy for human movements.

Accurately forecasting the spatial-temporal dynamics of infectious diseases requires considering mobility patterns and capturing local trends. Many models rely solely on historical daily case data for predictions \cite{wangpingExtendedSIRPrediction2020, ahouzPredictingIncidenceCOVID192021, zhaoPredictionGlobalOmicron2022}. However, this univariate analysis overlooks the spatial information and clinical outcomes of neighboring regions or countries, which can greatly aid in estimating local future cases \cite{franch-pardoSpatialAnalysisGIS2020}. Previous works utilizing the neural networks, featuring a temporal backbone and a GCN module, present a promising alternative by explicitly capturing associations in multivariate time series data \cite{dengColaGNNCrosslocationAttention2020, gaoSTANSpatiotemporalAttention2021, jinInterseriesAttentionModel2021, gaoEvidencedrivenSpatiotemporalCOVID192023}. This innovative approach effectively models the spatial-temporal structure of SARS-CoV-2 incidence, leading to more accurate forecasting than conventional epidemiological tools.

To enhance the representation of human movements with neural networks, we suggest assessing the influence of mobility data on forecasting tasks. In our pilot study, we compare the performance of models that only use incidence data with those that incorporate both incidence and mobility information. Recognizing the inherent challenges of missing data and geographical coverage limitations in mobility data, we advocate for alternative data sources to better capture mobility in forecasting tasks.

The relationship among clinical indicators—incidence, hospitalization rate, and mortality—is mutually dependent, as depicted in Supplementary Fig.2. Inspired by Multi-task Learning (MTL) \cite{caruanaMultitaskLearning1997}, which improves generalization by sharing representations among related tasks to mitigate data sparsity and overfitting risks \cite{zhangSurveyMultitaskLearning2022}, we suggest that incorporating mortality and hospitalization rates with incidence data can help the model infer mobility information from clinical data rather than relying solely on mobility data.

Beyond data, adjusting model structures offers another pathway for improving forecasting results. In our main experiments, we aim to evaluate the contribution of the GCN module in capturing human movement without explicitly relying on mobility data. Previous studies have explored various temporal backbones, such as LSTM \cite{wardForecastingSARSCoV2Transmission2022, gaoEvidencedrivenSpatiotemporalCOVID192023}, GRU \cite{kavourasCOVID19SpatioTemporalEvolution2022}, CNN \cite{jinInterseriesAttentionModel2021}, SIR models \cite{gaoSTANSpatiotemporalAttention2021}, and GCN modules, to represent spatial information during the pandemic. However, due to differences introduced by temporal structures, the effectiveness of the GCN module remains inadequately assessed. Therefore, we opt for the powerful transformer as the temporal backbone to evaluate the effectiveness of incorporating a GCN module with physical-distance-based and dynamically derived adjacency matrices.

To study the influence of cross-border commuting, we focus on two distinct regions: the European Union (EU), representing countries with interconnected mobility, and the United States (US), reflecting domestic mobility patterns. For our analysis, we have chosen 28 EU countries and 49 US states based on data availability. These regions serve as external validation for each other in developing a forecasting model for SARS-CoV-2. 

Our study aims to identify the primary contributor to enhancing forecasting capabilities by aiding in mobility retrieval. We propose using a transformer backbone to model temporal correlation, while spatial correlation is captured by a GCN module. The adjacency matrices include a physical-distance-based matrix generated using coordinates and a non-distance-related one generated by an attention mechanism with a normal truncating constraint. Additionally, by incorporating supplementary data across sub-regions, we test the contribution of MTL in improving predictive performance. A sensitivity analysis is carried out, comparing the prediction performance of the transformer backbone with that of the ARIMA model for univariate forecasting and tree-based and linear models for multivariate forecasting.

\begin{figure}[h]
\centering
\includegraphics[width=1.0\linewidth]{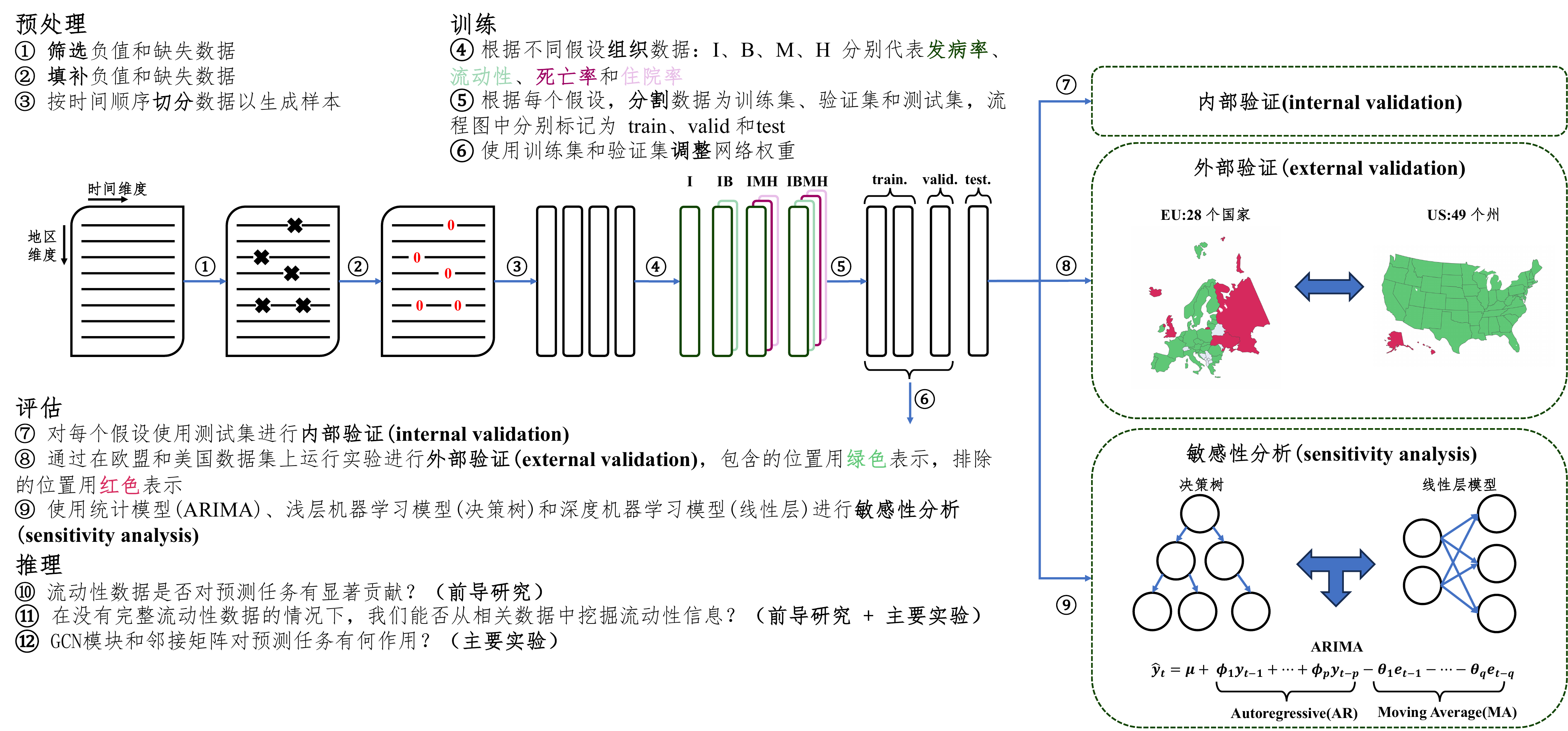}
\caption{\textbf{Research Pipeline} The project consists of four major steps, termed ``preprocess'', ``train'', ``evaluate'' and ``infer''. The first three steps serve as pipelines to answer the questions posed in the ``infer'' step.}
\label{fig:pipeline}
\end{figure}

\section*{Results}
\subsection*{Research Pipeline}
Our study aims to find the core contributor to the spatiotemporal neural networks for integrating mobility information into SARS-CoV-2 incidence forecasting. This objective is divided into three questions, outlined in the ``Infer'' section of Fig.~\ref{fig:pipeline}. The first question examines the necessity of mobility data in the incidence forecasting task, which will be addressed in the pilot study. This initial investigation will provide crucial insights to support subsequent experiments. Both the pilot study and the main experiment will delve into the possibility of extracting mobility information from related data. Building upon the findings of the preliminary inquiries, the main experiment will focus on exploring the effectiveness of the GCN module and the adjacency matrices for the forecasting task.

We denote the historical incidence as $\mathbf{X} = [\mathbf{x}_1, \dots, \mathbf{x}_T] \in \mathbb{R}^{N \times T}$ where $N$ and $T$ denote the number of sites and time window span. The objective is to predict the incidence rate $\hat{\mathbf{Y}} = [\hat{\mathbf{x}}_{T+1}, \dots, \hat{\mathbf{x}}_{T+1+F}] \in \mathbb{R}^{N \times F}$ for $F$ time steps ahead in the future. We formulate the COVID-19 epidemic prediction problem as follows:
\begin{align}
    [\mathbf{x}_1, \dots, \mathbf{x}_T] 
    \xrightarrow{f} 
    [ \hat{\mathbf{x}}_{T+1}, \dots, \hat{\mathbf{x}}_{T+1+F} ].
\end{align}

Our goal is to learn a mapping $f$ that accurately predicts the incidence rate $\mathbf{Y} = [\mathbf{x}_T, \dots, \mathbf{x}_{T+F}] \in \mathbb{R}^{N \times F}$ by minimizing the discrepancy between the predicted and actual values. The Mean Absolute Error (MAE) and Root Mean Squared Error (RMSE) are employed as evaluation metrics to quantify the difference between the predicted and actual values, i.e.,  
\begin{align}
    \text{MAE} &= \frac{1}{U}\sum_{j=1}^{U}|\hat{\mathbf{Y}
    }_j - \mathbf{Y}_j|, \nonumber \\
    \text{RMSE} &= \sqrt{\frac{1}{U}\sum_{j=1}^{U}(\hat{
    \mathbf{Y}
    }_j - \mathbf{Y}_j)^2}. \label{eq:loss_fn}
\end{align}

Four types of model structures are explored primarily in our study, including the vanilla Transformer (Trans), Transformer integrated with the GCN module using an adjacency matrix based on physical distances (Trans + GCN), Transformer integrated with the GCN module using an adaptively generated adjacency matrix (Trans + Adp), and Transformer integrated with the GCN module using both types of adjacency matrices (Trans + GCN + Adp). We adopt incidence (I), mortality (M), hospitalization rate (H), and Google mobility (B) as data sources in our study.  Mobility is defined as the variation in visits to locations, while the incidence, mortality, and hospitalization rate are defined as the ratio of documented case counts in raw data to the population per 10,000 individuals within subregions for the given year. This normalization ensures the stability of back-propagation during training. We train all models five times with different random initializations. Results are reported with the format $\mu \pm \sigma$ where $\mu$ and $\sigma$ represent the mean and standard deviations, respectively. 

The datasets used in both the pilot study and the main experiment undergo preprocessing to address missing and mislabeled data. Detailed information on the imputation process is provided in Supplementary Table 2. Mobility data often contains a significant amount of missing information compared to the incidence, mortality, and hospitalization rates data, as shown in Supplementary Table 3. Therefore, for the pilot study, we exclusively select mobility data from the workplace, from February 15, 2020, to November 15, 2022, to mitigate missing data at the beginning and end of the pandemic. For the main experiments, the IMH dataset from January 22, 2020, to November 27, 2022, is utilized. To demonstrate the model's robustness under data scarcity and distribution shift, we adopt the K-fold progressive cross-validation approach with $K = 5$, as depicted in Supplementary Fig.4. This approach ensures that the model remains blind to future observations.

To test our hypotheses, datasets are constructed by integrating various input sources. Each dataset is partitioned into chronologically ordered samples, which are then divided into training, validation, and test sets in a $7:1:2$ ratio. The training and validation sets are utilized to fine-tune and select the most effective model architecture. The test set, unseen during training, serves for internal validation. External validation is conducted by testing hypotheses using data from two distinct regions, the US and the EU. This ensures the generalizability of our findings across different geographical contexts. To assess the efficacy of our network, we compare its performance with that of statistical (ARIMA), shallow (Decision Trees), and deep (Linear layers) machine learning models in our sensitivity analysis.

\subsection*{Pilot Study: Assessing the Impact of Mobility Data on Forecasting Task}
\begin{figure}[ht!]
     \centering
     \begin{subfigure}[b]{0.45\textwidth}
         \centering
         \includegraphics[width=\textwidth]{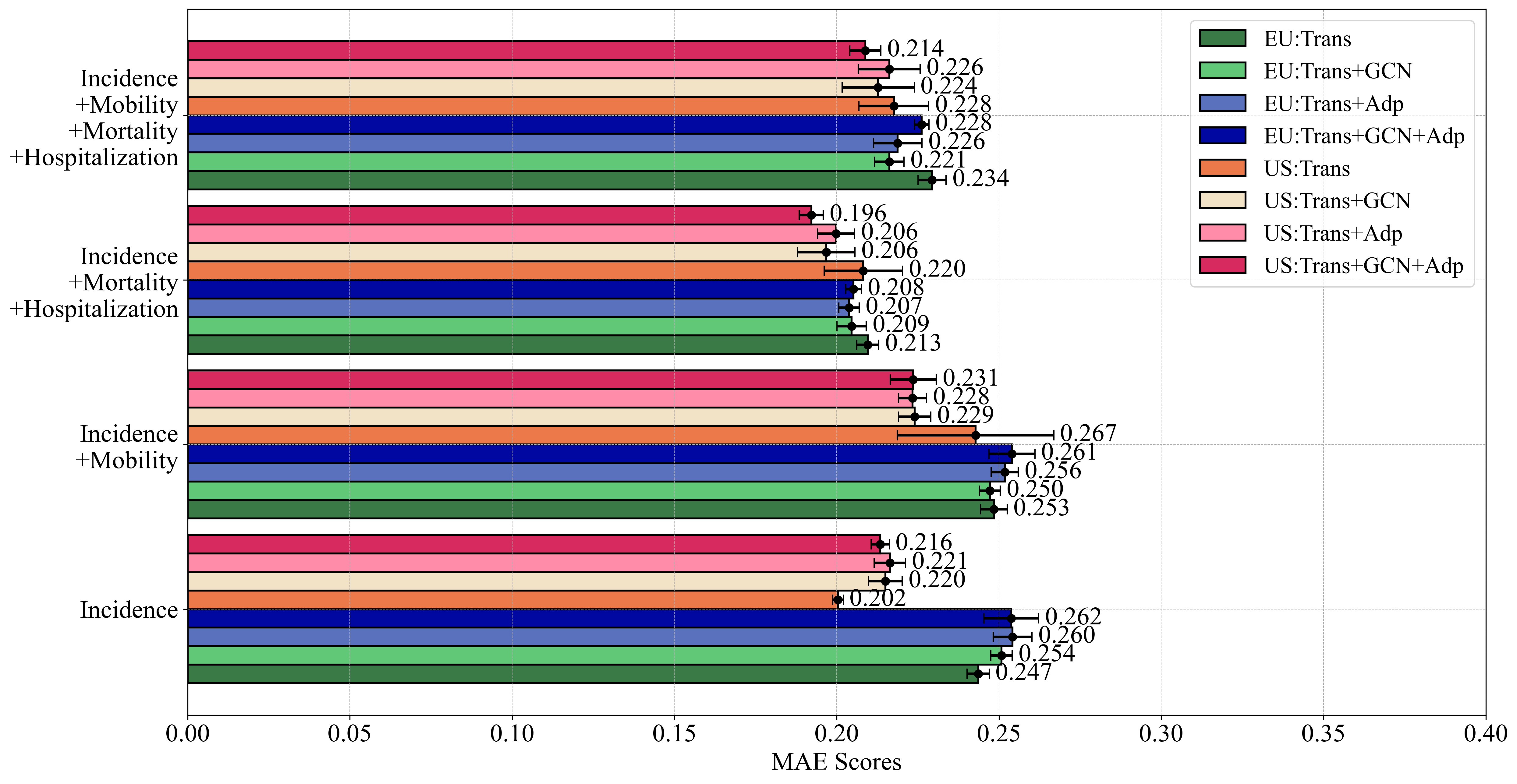}
         \caption{MAE scores}
         \label{fig:mae_pilot}
     \end{subfigure}
     \hfill
     \begin{subfigure}[b]{0.45\textwidth}
         \centering
         \includegraphics[width=\textwidth]{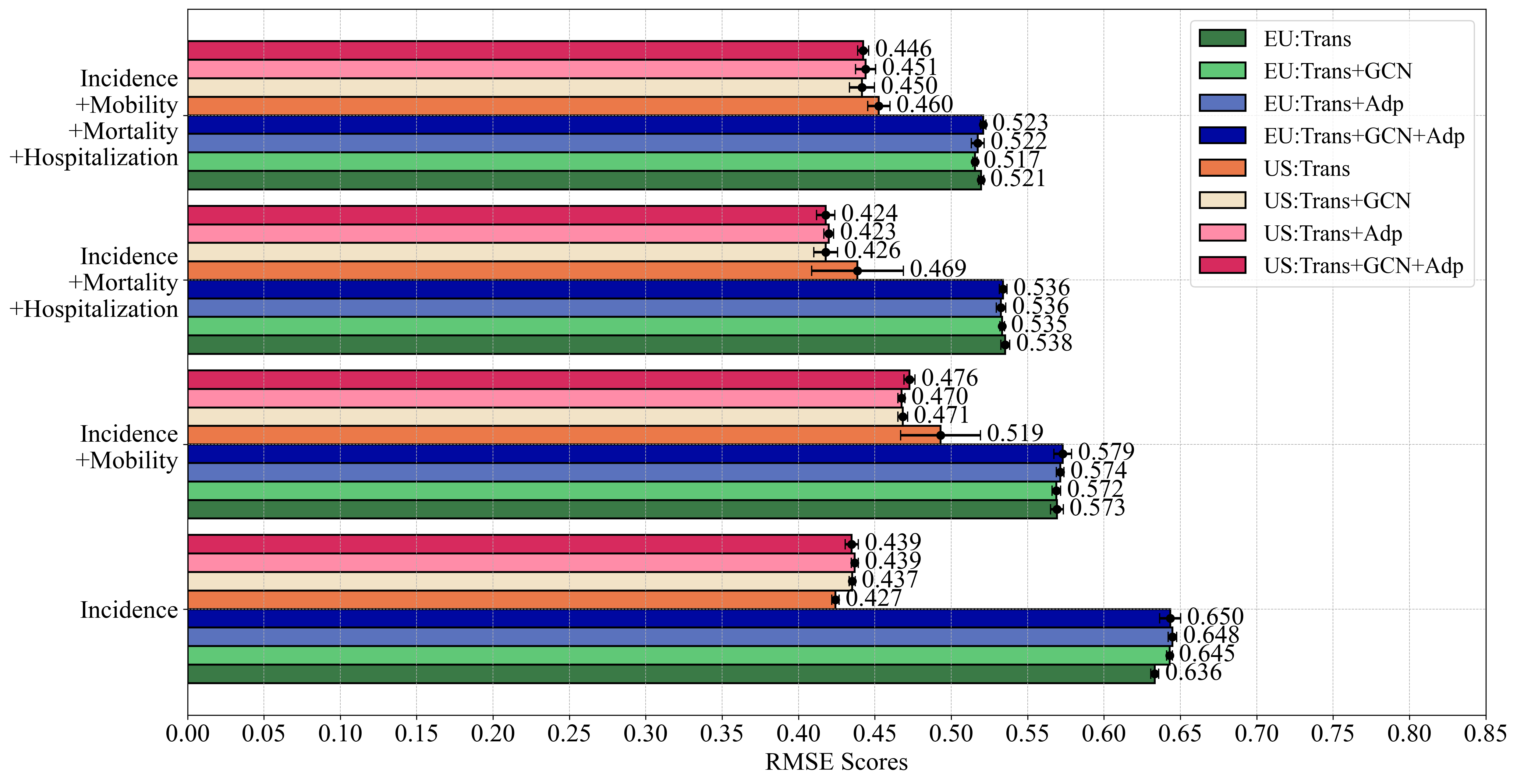}
         \caption{RMSE scores}
         \label{fig:rmse_pilot}
     \end{subfigure}
     \caption{\textbf{MAE and RMSE scores for incidence forecasting during the pilot study, spanning from July 2, 2022, to November 15, 2022.} The correspondence between input and output lengths is maintained at 12 steps each.}
     \label{fig:pilot_mae_rmse}
\end{figure}

In our preliminary investigation, we aim to address the question: ``Does mobility data significantly contribute to the forecasting task?'' As depicted in Fig.~\ref{fig:pilot_mae_rmse}, we present the performance in MAE and RMSE scores for four different types of input data, arranged from bottom to top on the vertical axis. 

Our fundamental null hypothesis posits that mobility data has no discernible contribution to the forecasting task. By comparing models using ``Incidence+Mobility'' (IB) with ``Incidence'' (I) as input data, we establish a fundamental alternative hypothesis suggesting that mobility data enhances forecasting accuracy and, thus, is a good proxy for human movement information. Additionally, by comparing ``Incidence+Mortality+Hospitalization'' (IMH) with ``Incidence+Mobility'', we propose an alternative hypothesis indicating that mobility information can be extracted from related data, such as mortality and hospitalization data. Further, by comparing ``Incidence+Mobility+Mortality+Hospitalization'' (IBMH) with "Incidence+Mortality+Hospitalization", we propose an alternative hypothesis suggesting that mortality and hospitalization data alone are not adequate sources of mobility information, which implies that the incorporation of mobility data is not redundant for accurate forecasting.

Supplementary Table 1 presents the P-values from the one-sided student t-test for these hypotheses. Our analysis indicates that mobility data does not significantly contribute to the forecasting task, with only the test in the EU RMSE scores supporting its effectiveness. Thus, mobility data may not serve as a good proxy for human movement information for forecasting tasks using neural networks. On the contrary, results have shown that mobility information can be extracted from mortality and hospitalization data, as indicated by significantly smaller MAE and RMSE scores for both regions when using IMH instead of IB or I. Moreover, mortality and hospitalization data alone are deemed sufficient sources of mobility information, with the IMH dataset demonstrating the best performance across all model structures and regions. Comparing IBMH with IMH results, no significant improvement is detected in terms of MAE scores for both regions.

\begin{figure}[h]
     \centering
     \begin{subfigure}[b]{\textwidth}
         \centering
         \includegraphics[width=\textwidth]{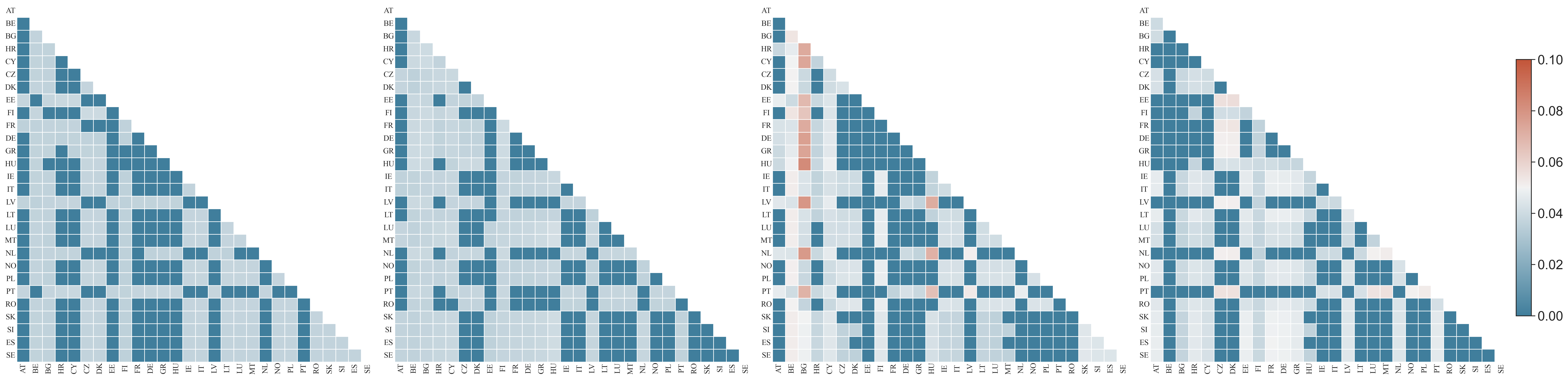}
         \caption{EU dynamic adjacency matrix}
         \label{fig:eu_matrix_pilot_trans_adp}
     \end{subfigure}
     

     \hfill
     \begin{subfigure}[b]{\textwidth}
         \centering
         \includegraphics[width=\textwidth]{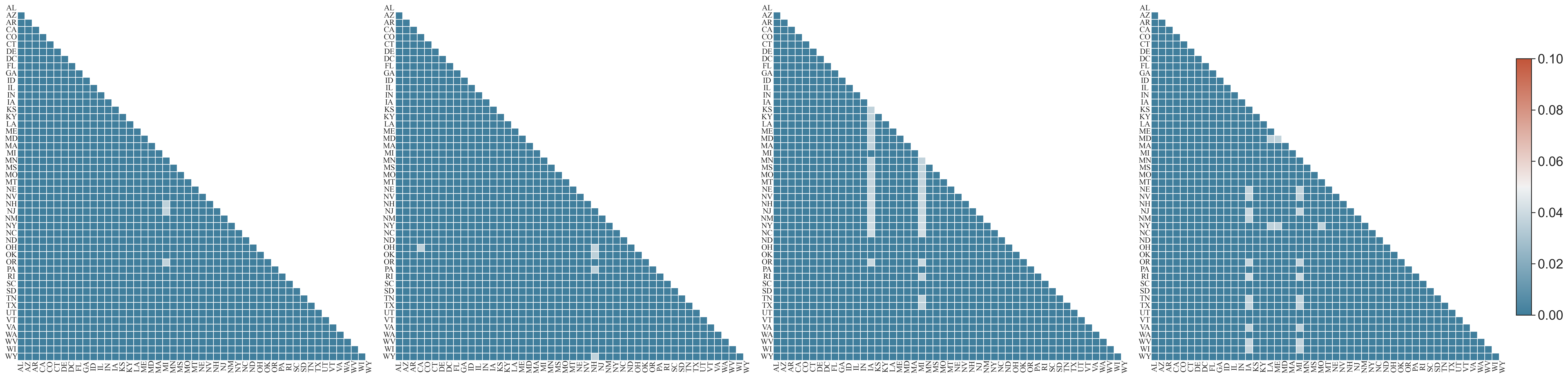}
         \caption{US dynamic adjacency matrix}
         \label{fig:us_matrix_pilot_trans_adp}
     \end{subfigure}
     
     \caption{\textbf{Heatmaps of ``Trans+Adp'' dynamic maps from the pilot datasets of the EU and US.} Moving from left to right, the datasets used are I, IB, IMH, and IBMH.}
     \label{fig:pilot_matrix}
\end{figure}

One of the central designs in the spatiotemporal networks involves dynamically generating adjacency matrices to represent the changing nature of mobility over time. These matrices are visually represented using heatmaps, as depicted in Fig.~\ref{fig:pilot_matrix}. In these heatmaps, weights between locations are color-coded, with red indicating strong connections and blue indicating weaker ones. Comparing the rows, we observe that US maps are more sparse and uniform than EU maps. Comparing the columns, we discover that upon adding mobility data, both the EU and US maps exhibit similar uniformity. Notably, no important connections become more evident in the second column compared to the first. Furthermore, comparing the third column to the second, we observe varied effects upon incorporating mortality and hospitalization data in both the EU and US datasets. In the EU dataset, strong weights are evident, highlighted by conspicuous red patterns. However, such red patterns are absent in the US dataset. Finally, the last column, incorporating all data sources, shows a more uniform map for both regions compared to the third column. This suggests that adding mobility to mortality/hospitalization data does not help in capturing important connections.

Our findings offer evidence supporting two key conclusions. First, mobility data does not significantly contribute to the forecasting task, as not enough evidence shows improved predictive performance when incorporating mobility data into the models. Second, mobility information can be effectively mined from related data, such as mortality and hospitalization data, suggesting that these data sources contain valuable insights into mobility patterns that can enhance forecasting accuracy.

\begin{figure}[h]
    \centering
    \includegraphics[width=\textwidth]{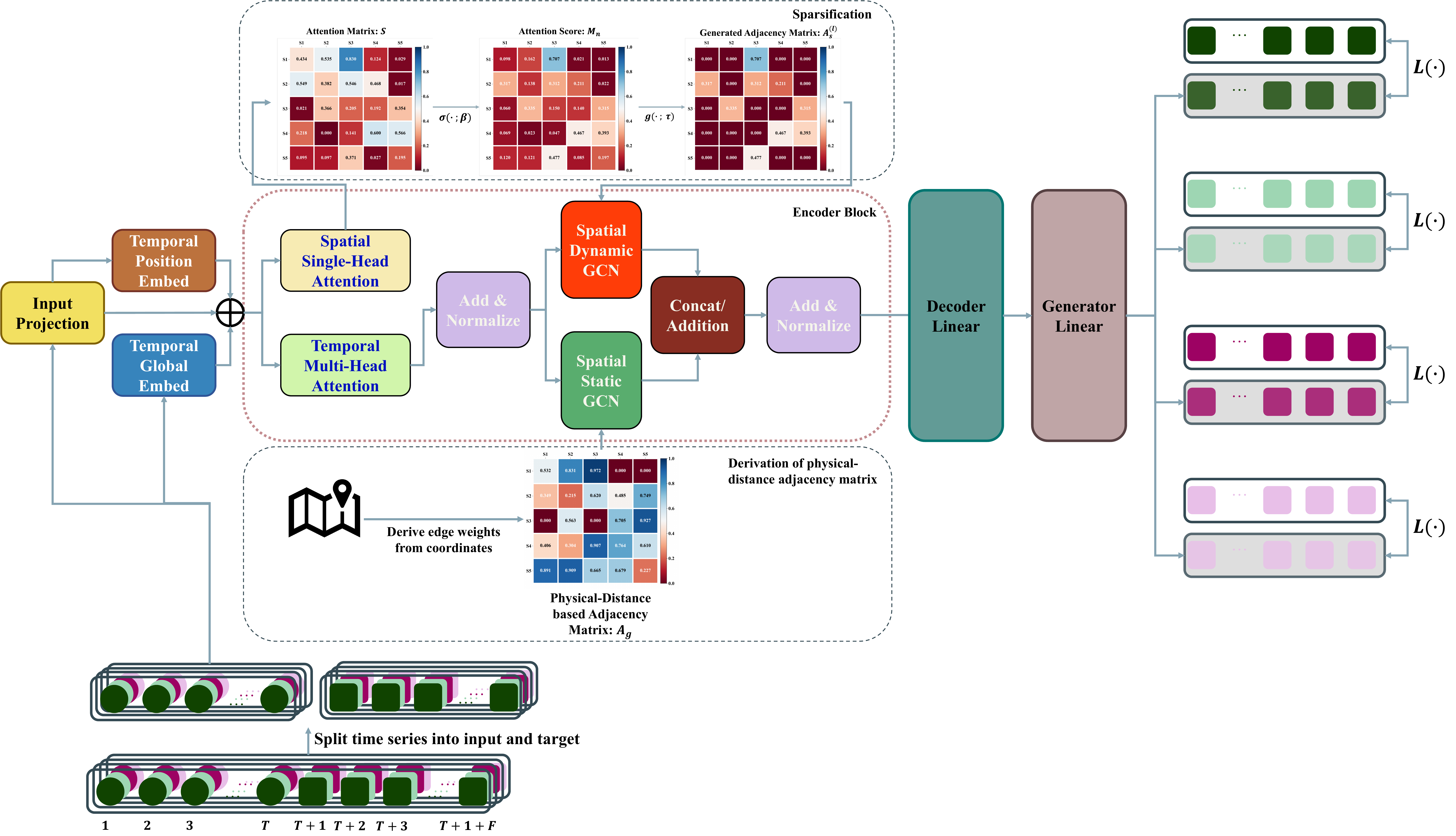}
    \caption{\textbf{Network Structure of ``Trans+GCN+Adp''}}
    \label{fig:network}
\end{figure}

\begin{figure}[h!]
     \centering
     \begin{subfigure}[b]{0.96\textwidth}
         \centering
         \includegraphics[width=1\linewidth]{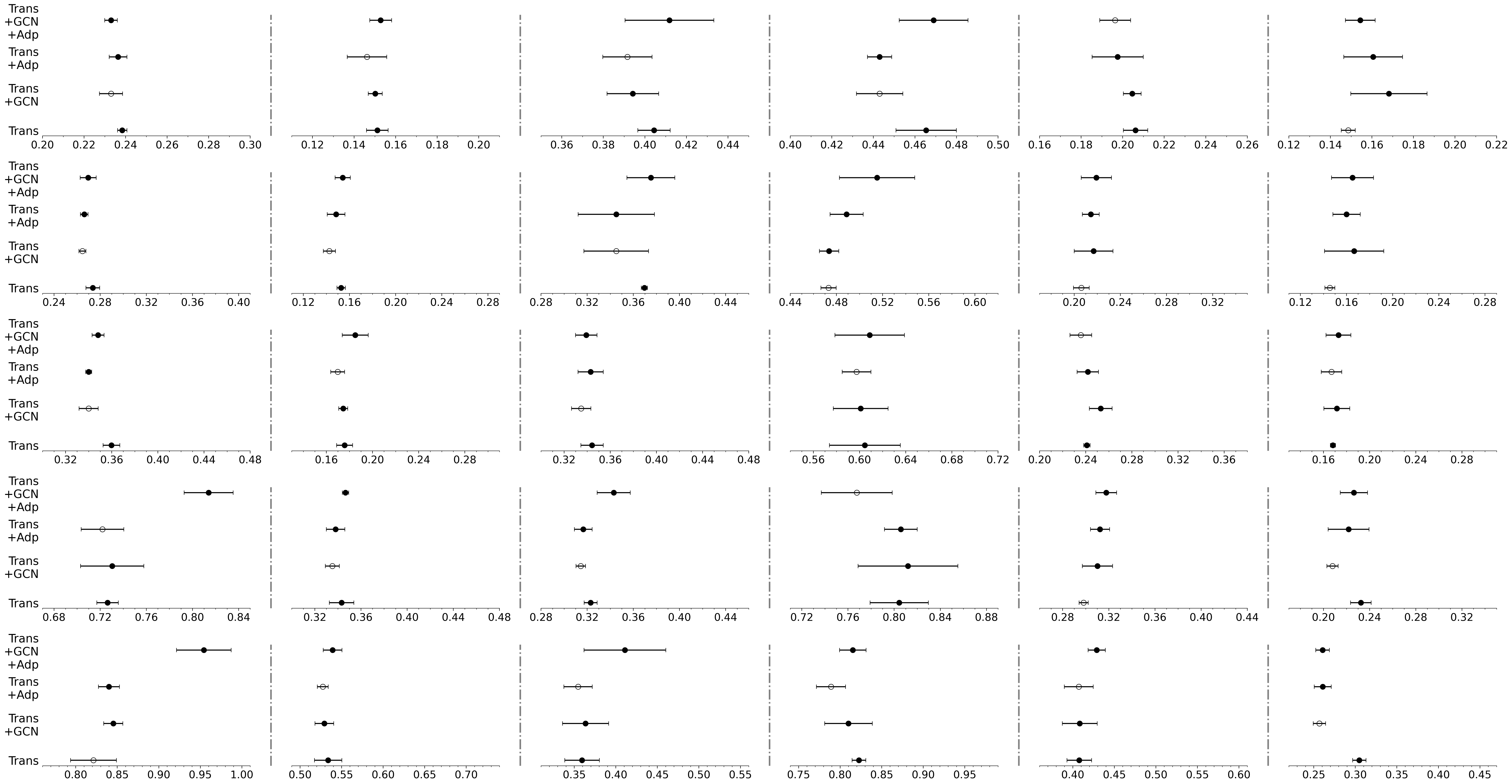}
         \caption{EU MAE Scores}
         \label{fig:eu_mae_testset}
     \end{subfigure}
     \hfill
     \begin{subfigure}[b]{0.96\textwidth}
         \centering
         \includegraphics[width=1\linewidth]{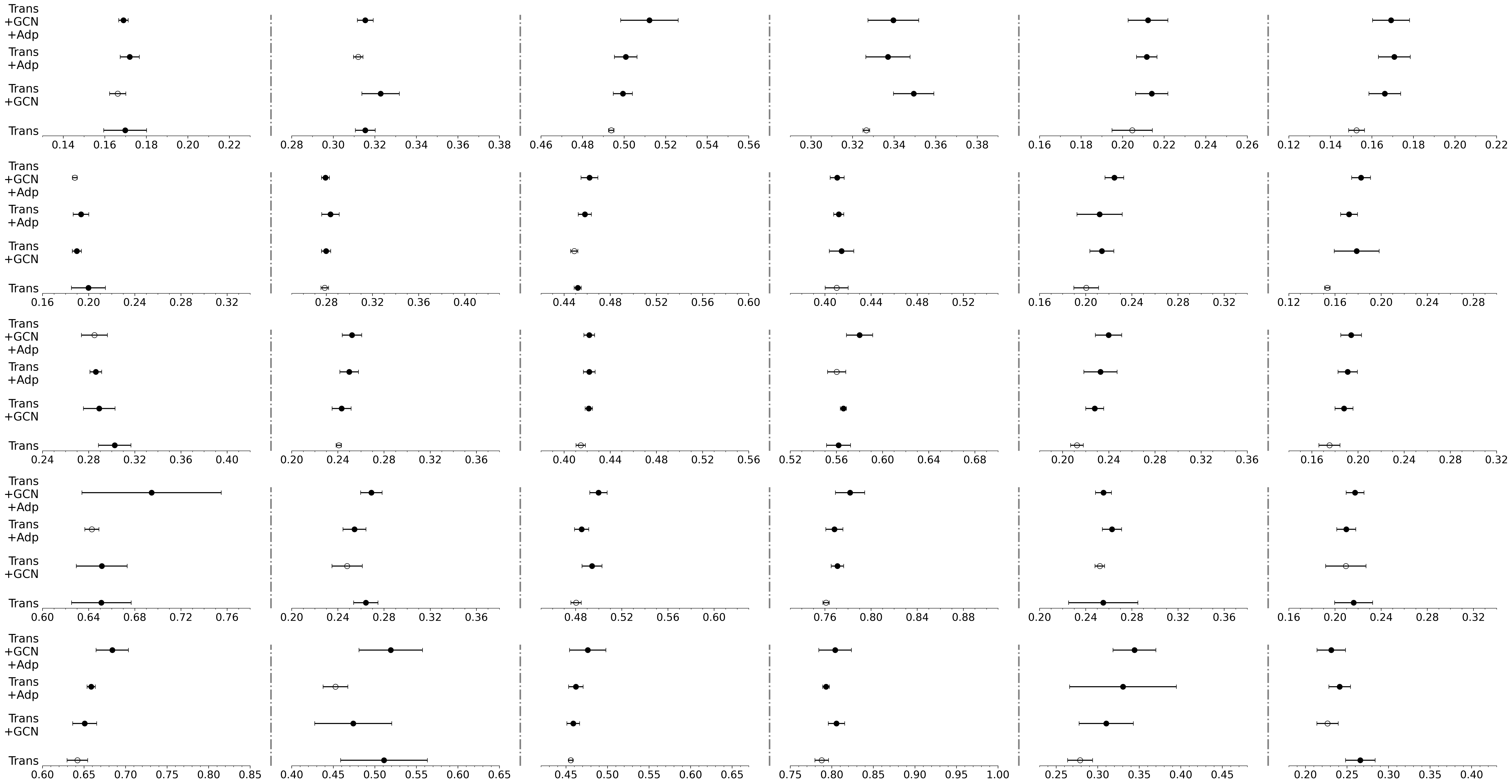}
         \caption{US MAE Scores}
         \label{fig:us_mae_testset}
     \end{subfigure}
     \caption{\textbf{Incidence forecasting MAE scores for test sets of 5-fold split and the final test set.} The rows represent the output lengths, $3, 6, 12, 24, 36$. The columns represent folds.}
     \label{fig:test_set_mae}
\end{figure}

\begin{figure}[h!]
     \centering
     \begin{subfigure}[b]{0.96\textwidth}
         \centering
         \includegraphics[width=1\linewidth]{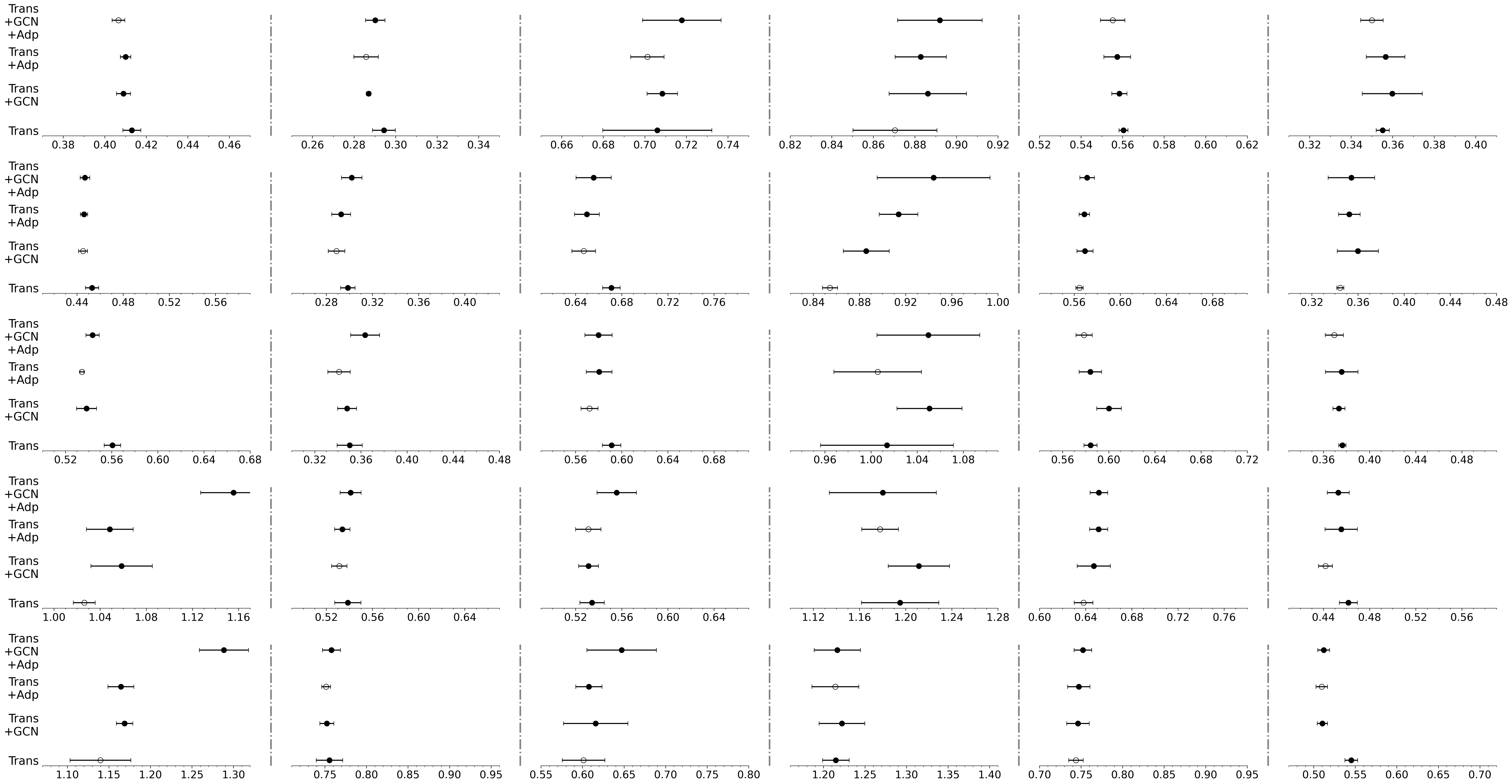}
         \caption{EU RMSE Scores}
         \label{fig:eu_rmse_testset}
     \end{subfigure}
     \hfill
     \begin{subfigure}[b]{0.96\textwidth}
         \centering
         \includegraphics[width=1\linewidth]{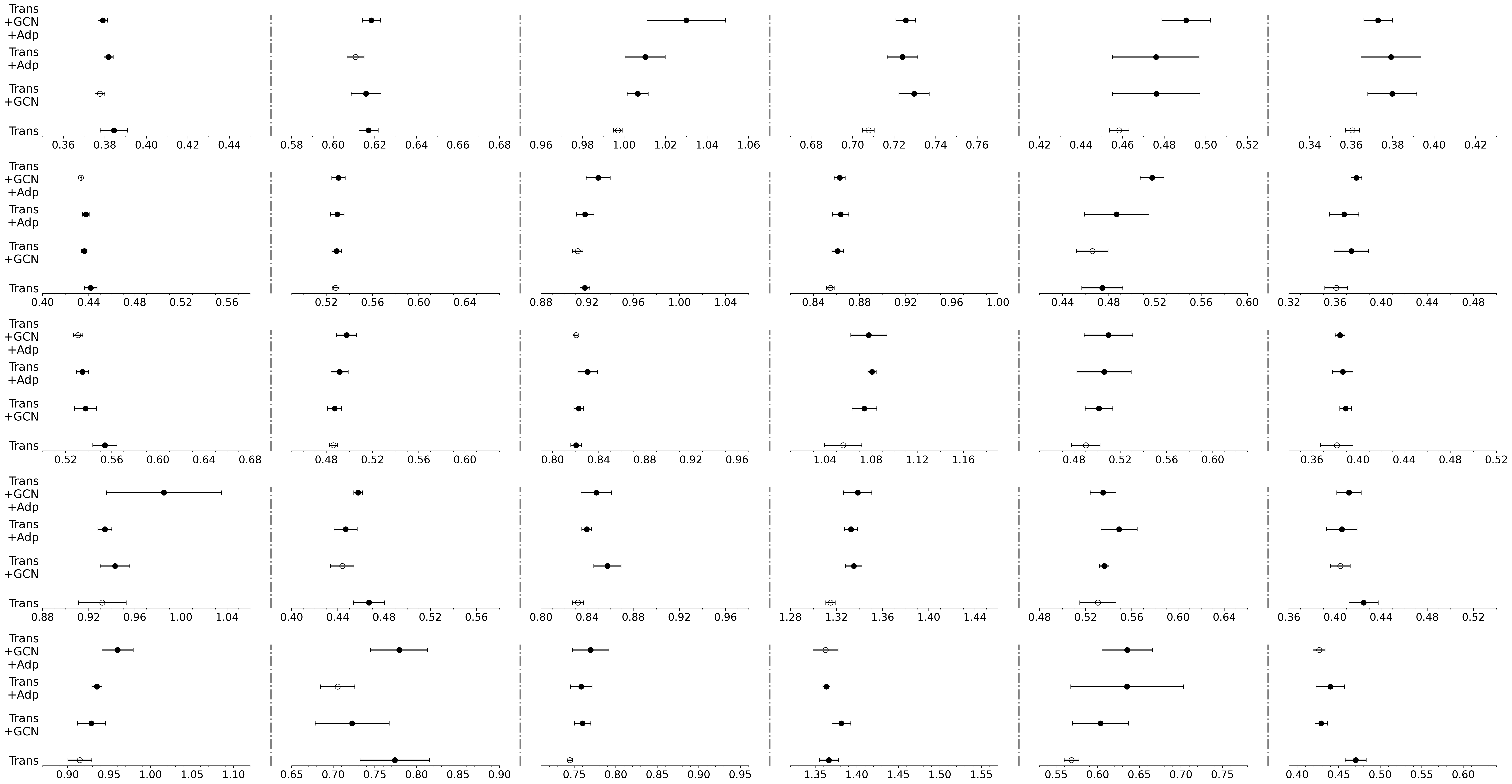}
         \caption{US RMSE Scores}
         \label{fig:us_rmse_testset}
     \end{subfigure}
     \caption{\textbf{Incidence forecasting RMSE scores for test sets of 5-fold split and the final test set.} The rows represent the output lengths, $3, 6, 12, 24, 36$. The columns represent folds.}
     \label{fig:test_set_rmse}
\end{figure}

\subsection*{Main experiments: Determining the Optimal Network Structure for Mobility Information Retrieval}

We design experiments to answer the following research questions:
\begin{enumerate}
    \item How much does the GCN module contribute to the forecasting task?
    \item How does the GCN module performance vary under different lengths of prediction window?
    \item How does the GCN module performance vary under data scarcity and data distribution shift?
    \item What are the differences and similarities in prediction between EU and US regions?
    \item Do dynamic maps accurately reflect the impact of human movement restrictions imposed by lockdown orders?
\end{enumerate}

\subsubsection*{Contribution of the GCN module}
We test the contribution of the GCN module with the static adjacency matrix (Trans+GCN), the dynamic adjacency matrix (Trans+Adp), and both matrices (Trans+GCN+Adp) under the 5-fold split strategy for different output lengths. The results are summarized in Fig.~\ref{fig:test_set_mae} and Fig.~\ref{fig:test_set_rmse}. Using a one-sided t-test with $\alpha=0.05$, similar to the pilot study, we test the significance of model improvements.

From the fold split perspective, comparing Trans+Adp with Trans shows an increase in MAE scores of $0.08\%$, $1.18\%$, $0.00\%$, $-0.30\%$, and $-1.20\%$, failing the significance test with $P=0.504, 0.561, 0.500, 0.487$, and $0.434$. In RMSE scores, the fold average for output lengths $12$ and $36$ show slight improvements $(0.18\%, P=0.490$, $0.07\%, P=0.496)$ without statistical significance. Comparing Trans+GCN with Trans, we find an increase in MAE score by $1.04\%$, $0.60\%$, $0.25\%$, $-0.37\%$ and $-0.84\%$, but none of the improvements proving significant with $P=0.552, 0.531, 0.512, 0.484, 0.455$. Comparing Trans+GCN+Adp with Trans, we find a deteriorated performance that MAE has increased by $1.77\%$, $3.94\%$, $1.63\%$, $3.32\%$ and $4.46\%$, which are also not significant $P=0.585, 0.692, 0.578, 0.641$ and $0.722$.

For six output lengths, Trans+Adp shows an average MAE increase over the vanilla transformer by $-0.62\%$, $-2.85\%$, $-0.44\%$, $-0.39\%$, $4.47\% $, and $-0.40\%$, which were not significant with $P=0.477, 0.369, 0.442, 0.473, 0.800$, and $0.467$. RMSE scores for fold 1 and the final test set showed improvements but failed the significance tests ($2.36\%, P=0.354$, $1.27\%, P=0.326$).

On average, across folds, output lengths, and regions, Trans+GCN, Trans+Adp, and Trans+GCN+Adp do not significantly decrease MAE scores over the vanilla transformer model, with $0.04\%, 0.22\%$, and $-3.21\% (P=0.497, 0.479, 0.767)$. No decrease is observed for the GCN module in terms of RMSE scores. Therefore, we conclude that, on average, the addition of the GCN module with either or both adjacency matrices lacks evidence to prove their contribution to forecasting performance.

\subsubsection*{Perfomrance analysis for different lengths}
In Fig.~\ref{fig:test_set_mae} and Fig.~\ref{fig:test_set_rmse}, MAE and RMSE scores for test sets are depicted, respectively. Each row represents one output length ($3, 6, 12, 24$, and $36$), illustrating short-term and long-term prediction scenarios. For the EU dataset, adding dynamic maps shows performance improvement supported by MAE ($-2.40\%, -1.83\%, -2.18\%$, $P=0.418, 0.441, 0.417$) and RMSE ($-0.18\%, -1.59\%, -0.12\%$, $P=0.493, 0.436, 0.494$) scores for output lengths $3, 12$, and $36$, without statistical significance. MAE scores decrease uniformly for adding static maps, but RMSE scores do not consent. For the US dataset, decreases are observed in MAE ($-0.12$, $P=0.495$) and RMSE ($-0.02$, $P=0.499$) scores for output length $36$ when comparing Trans+Adp to the Trans model, but no significance is found. Trans+GCN shows similar minor improvements without significance. Pooling results from both regions, agreement between MAE ($-1.20\%$, $P=0.434$) and RMSE ($-0.07\%$, $P=0.496$) scores is observed in adding dynamic maps to the transformer for output length $36$, but the improvement fails the significance test.

\subsubsection*{Performance analysis under varying data availability}
In Fig.~\ref{fig:test_set_mae} and Fig.~\ref{fig:test_set_rmse}, the performance for fold split and the final test sets are displayed as columns, representing fold $1, 2, 3, 4, 5$ and the final test set from left to right. These columns aim to demonstrate model robustness against data scarcity and distribution shifts. For the EU dataset, Trans+Adp's performance improvement is evident in MAE ($-1.99\%, -2.75\%, -2.94\%$, $P=0.450, 0.128, 0.351$) and RMSE ($-1.45\%, -1.08\%, -1.59\%$, $P=0.450, 0.353, 0.372$) scores for fold $2, 3$, and the final test set, without statistical significance. Similar results are found when comparing Trans+GCN to Trans. Trans+GCN+Adp also shows decreases in MAE ($-2.00\%$, $P=0.397$) and RMSE ($-2.14\%$, $P=0.333$) in the final test set. For the US dataset, both Trans+Adp and Trans+GCN demonstrate decreases in MAE (Trans+Adp: $-0.62\%, -3.58\%$, $P=0.484, 0.327$; Trans+GCN: $-0.91\%, -2.62\%$, $P=0.477, 0.379$) and RMSE (Trans+Adp: $-0.11\%, -3.07\%$, $P=0.496, 0.279$; Trans+GCN: $-0.13\%, -2.57\%$, $P=0.495, 0.319$) scores without statistical significance for fold 1 and 2. Pooling results from both regions, there is an agreement between MAE (Trans+Adp: $-2.85\%, -0.40\%$, $P=0.369, 0.467$; Trans+GCN: $-2.27\%, -1.18\%$, $P=0.397, 0.398$) and RMSE (Trans+Adp: $-2.36\%, -1.27\%$, $P=0.354, 0.326$; Trans+GCN: $-2.03\%, -1.46\%$, $P=0.374, 0.297$) scores in adding dynamic and static maps to the transformer for fold 2 and the final test set, but the improvement fail the significance test.

\subsubsection*{Comparative analysis between EU and US regions}
To compare the EU and US results, we average the results per row and column in Fig.~\ref{fig:eu_mae_testset} and Fig.~\ref{fig:eu_rmse_testset} as EU MAE and EU RMSE. Results in Fig.~\ref{fig:us_mae_testset} and Fig.~\ref{fig:us_rmse_testset} are averaged to obtain US MAE and US RMSE. Testing the EU and US datasets separately, we did not find sufficient evidence to support the notion that adding the adjacency matrix and the GCN module improves the performance of the transformer model. With the static matrix, $-0.95\% (P=0.441)$ and $0.90\% (P=0.562)$ increases are observed in EU and US MAE scores. With the dynamic matrix, $-1.38\% (P=0.414)$ and $0.97\% (P=0.567)$ and with both matrices, $2.92\% (P=0.671)$ and $3.51\% (P=0.725)$ are seen for EU and US MAE respectively. Although none of these differences are statistically significant, it is worth noting that the EU datasets appear to be more sensitive to the GCN module, and there are slight improvements in both MAE and RMSE scores when dynamic maps are added.

Several factors contribute to this performance disparity, including data quality, subregional connectivity, and calculation bias. Firstly, data quality plays a crucial role, and it is influenced by the collection and organization process. In our study, we utilize state-level and county-level data for the US, while only country-level data is available for the EU. This difference in granularity suggests that the US data, being aggregated, may have fewer missing values compared to the EU data, as shown in Supplementary Table 4. Secondly, variations in subregional connectivity also contribute to the performance differences. Despite both regions having a federal structure, the response of the US federal government to lockdown orders was more ambivalent, and enforcement was less strict compared to the EU \cite{renPandemicLockdownTerritorial2020, plumperLockdownPoliciesDynamics2022}. This difference in enforcement leads to greater mobility between states in the US, which may impact the spatial connectivity within the network differently compared to the EU. Lastly, the discrepancy in the number of sites considered in each region may introduce biases. If fewer sites are included in one region, the estimation error in those sites would have a more significant impact on the final performance metric. Therefore, the importance of site selection and the potential influence of regional characteristics on model performance cannot be overlooked. Overall, these factors underscore the complexity of forecasting infectious diseases and highlight the need for careful consideration of data quality, regional characteristics, and connectivity patterns when developing forecasting models.

\subsubsection*{Correlation between dynamic maps and human movement restrictions}
The effectiveness of NPIs, such as school closures and stay-at-home orders, in mitigating the COVID-19 pandemic has been extensively studied \cite{braunerInferringEffectivenessGovernment2021, lisonEffectivenessAssessmentNonpharmaceutical2023}. The impact of specific NPIs varies, but overall, their implementation has shown positive effects by reducing mobility \cite{snoeijerMeasuringEffectNonPharmaceutical2021}. Research indicates that lifting stringent NPIs, like lockdown measures, requires coordinated efforts among closely connected countries to minimize the risk of resurgence \cite{ruktanonchaiAssessingImpactCoordinated2020a}, highlighting the influence of spatial connections on lockdown efficacy. 

Dynamic maps generated from the spatial attention mechanism provide an additional perspective on human movement beyond forecasting outcomes. We introduce the mobility indicator $\Pi$, defined as the ratio of non-zero elements in the dynamic maps. Specifically, the mobility indicator $\Pi$ is calculated using the formula:
\begin{align}
    \Pi = \sum_{j=1}^{N} \sum_{i=1}^{N} \frac{I(\mathbf{M}_{n, (i, j)})}{N^2},
\end{align}
where the Indicator function $I(\cdot)$ determines if the element from $i$-th row and $j$-th column in dynamic map $\mathbf{M}_n$, denoted as $\mathbf{M}_{n, (i, j)}$,  is positive weight. Human movement restrictions, such as social distancing and quarantine orders, are challenging to track. Therefore, we use the duration of mandatory lockdown orders as a proxy for human movement restrictions. We define the input sample with its first data point aligning with the starting date of the lockdown as the first input and with the ending date as the last input. Thus, the number of samples considered lockdown samples is the same as the number of days included in the lockdown orders. We employ an output length of $36$ due to the minor improvement observed with the GCN module. Dynamic maps from Trans+Adp and Trans+GCN+Adp are included in the following analysis.

\begin{figure}
    \centering
    \includegraphics[width=\textwidth]{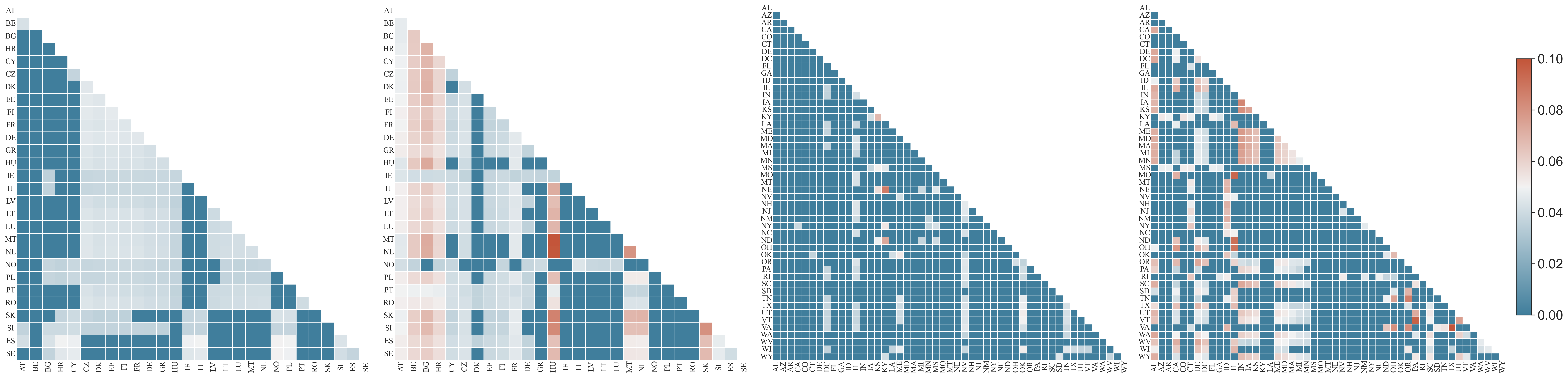}
    \caption{\textbf{Heatmaps of most connected dynamic maps.} Moving from left to right, EU Trans+Adp, EU Trans+GCN+Adp, US Trans+Adp, US Trans+GCN+Adp.}
    \label{fig:main_exp_matrix}
\end{figure}

Comparing the dynamic maps with the static ones, we find $\Pi=44.40\%, 45.66\%$ for EU maps and $\Pi=49.19\%, 9.48\%$ for US maps. The difference in connectivity between US static maps and dynamic maps can be attributed to the fact that by sharing the same truncating threshold with the EU, it is more difficult for US weights to be considered significant. However, this truncating threshold improved the forecasting performance for the US. Comparing the Trans+Adp with the Trans+GCN+Adp, we observe that the dynamic maps are slightly denser in EU Trans+Adp $\Pi=48.45\%, 40.34\%$ and US Trans+GCN+Adp $\Pi=5.29\%, 13.68\%$. For the EU Tran+Adp and Trans+GCN+Adp model, the highest connectivity is found for output length $3$ with $\Pi=51.53\%$ and $12$ with $\Pi=46.47\%$. Viewing from the fold perspective, we observe the highest connectivity $(\Pi=61.38\%)$ in fold 4 for output length $36$, and $(\Pi=69.29\%)$ in fold 2 for length $12$ as shown in Fig.~\ref{fig:main_exp_matrix}. For the US Trans+Adp and Trans+GCN+Adp models, the highest connectivity is found for output length $12$ with $\Pi=6.21\%$ and $12$ with $\Pi=19.53\%$. Viewing from the fold perspective, we observe the highest connectivity $(\Pi=15.72\%)$ in fold 1 for output length $24$, and $(\Pi=28.90\%)$ in fold 4 for output length $12$ as shown in Fig.~\ref{fig:main_exp_matrix}.

Since most lockdown orders occur early in the pandemic, we analyze all available maps, even if lockdown periods overlap with the training and validation sets. We calculate the mobility indicator $\Pi$ for pre-lockdown, lockdown, and post-lockdown periods as shown in Supplementary Table 2 and Table 3. Pre-lockdown samples span 24 days, while post-lockdown periods vary, with a 24-day span being typical, except for Luxembourg, where a 40-day span is used. The mobility indicator is derived from dynamic maps generated by Trans+Adp, with an output length of 36 for the final test set. A normal truncating threshold of 0.0357 and 0.0179 is applied for the EU and US, respectively. In the EU, the network captures mobility restrictions during periods with multiple subregions issuing lockdown orders, notably from mid-March to early May 2020 and late October to mid-December 2020. However, for countries with multiple lockdowns, waning indicator responses are observed, possibly due to pandemic fatigue. In the US, lockdown orders are concentrated between mid-March and mid-May 2020. While pre-lockdown and lockdown indicators show little change, post-lockdown indicators exhibit significant shifts, indicating a consistent rebound pattern after lockdowns.

\subsection*{Sensitivity Analysis}
To demonstrate the forecasting capabilities of shallow machine learning, statistical, and deep learning models, we perform a sensitivity analysis using decision trees, ARIMA, and DLinear. Boosting tree-based models are represented by XGBoost and LightGBM. Bagging tree-based models are represented by RandomForest. For statistical models, we employ a modified Hyndman-Khandakar algorithm with step-wise performance tuning. Since the statistical model is unable to handle the high-dimensional IMH dataset, it is excluded from the corresponding experiments. The linear layers-based deep learning model serves as a benchmark model, highlighting the impact of disregarding spatial correlation. Hyper-parameters are fine-tuned using the validation set. 

\begin{figure}[p]
     \centering
     \begin{subfigure}[b]{0.45\textwidth}
         \centering
         \includegraphics[width=\textwidth]{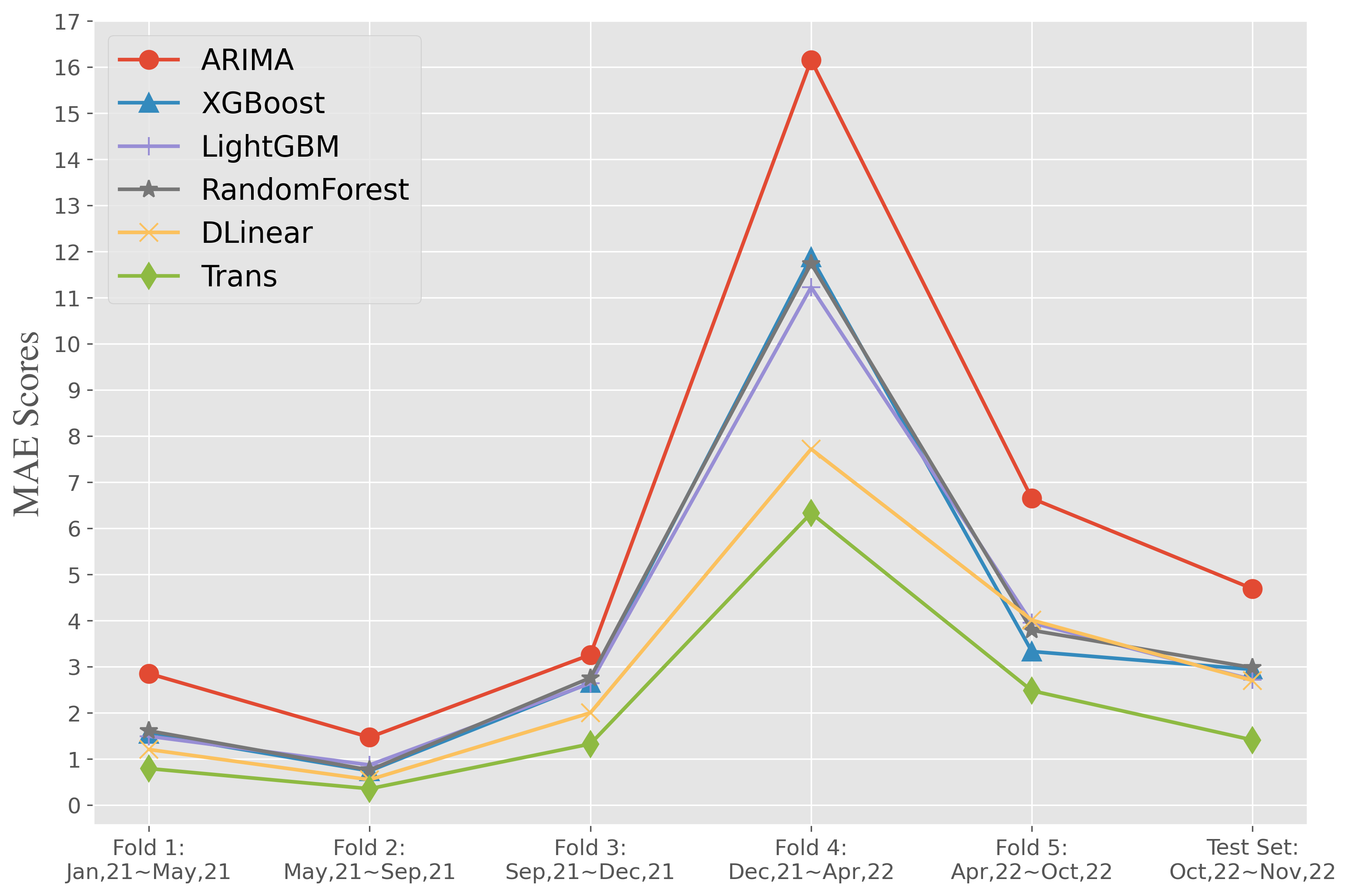}
         \caption{EU MAE scores using I}
         \label{fig:eu_mae_allfolds_nosupp}
     \end{subfigure}
     \hfill
     \begin{subfigure}[b]{0.45\textwidth}
         \centering
         \includegraphics[width=\textwidth]{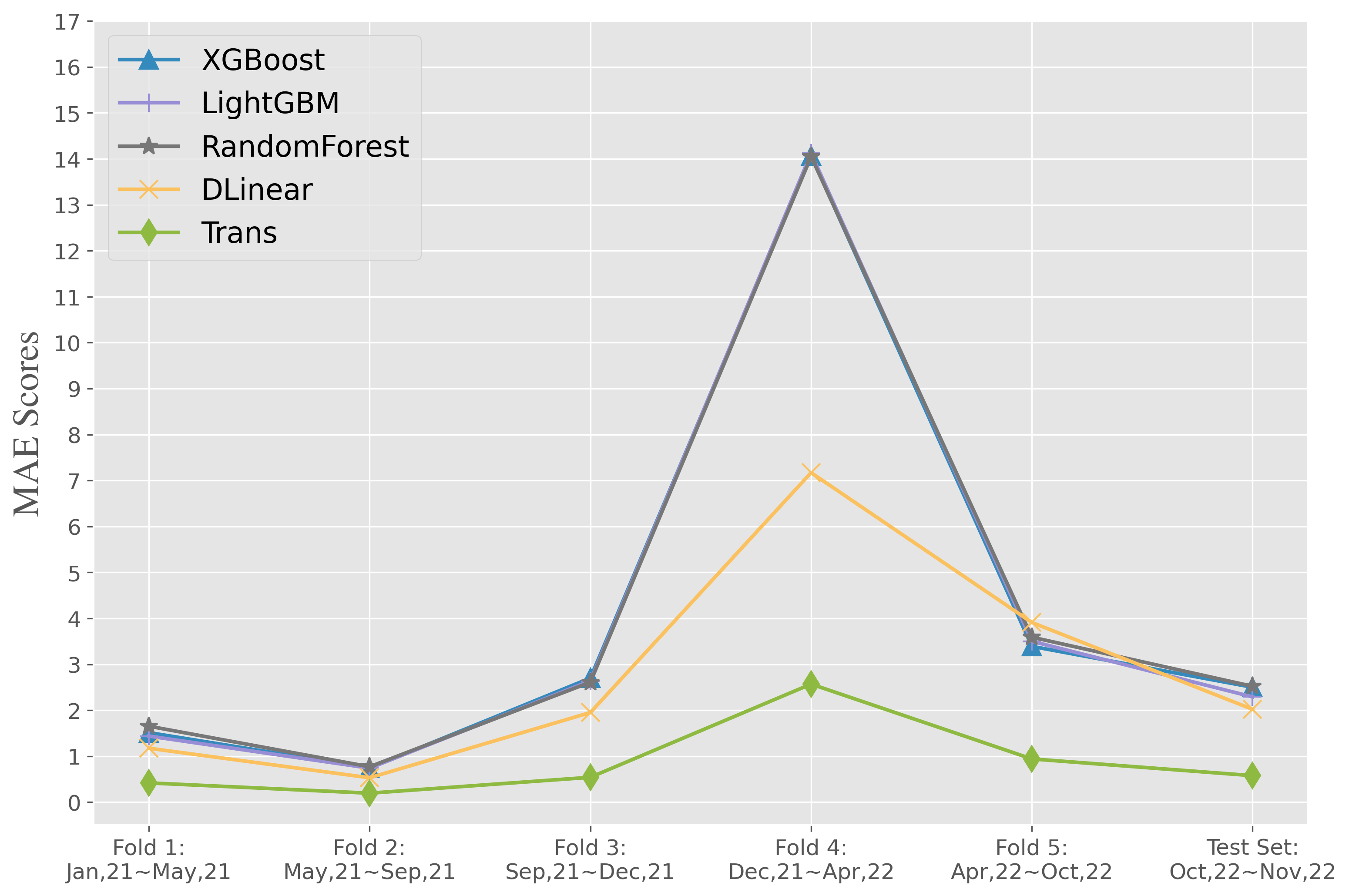}
         \caption{EU MAE scores using IMH}
         \label{fig:eu_mae_allfolds_supp}
     \end{subfigure}
     \hfill
     \begin{subfigure}[b]{0.45\textwidth}
         \centering
         \includegraphics[width=\textwidth]{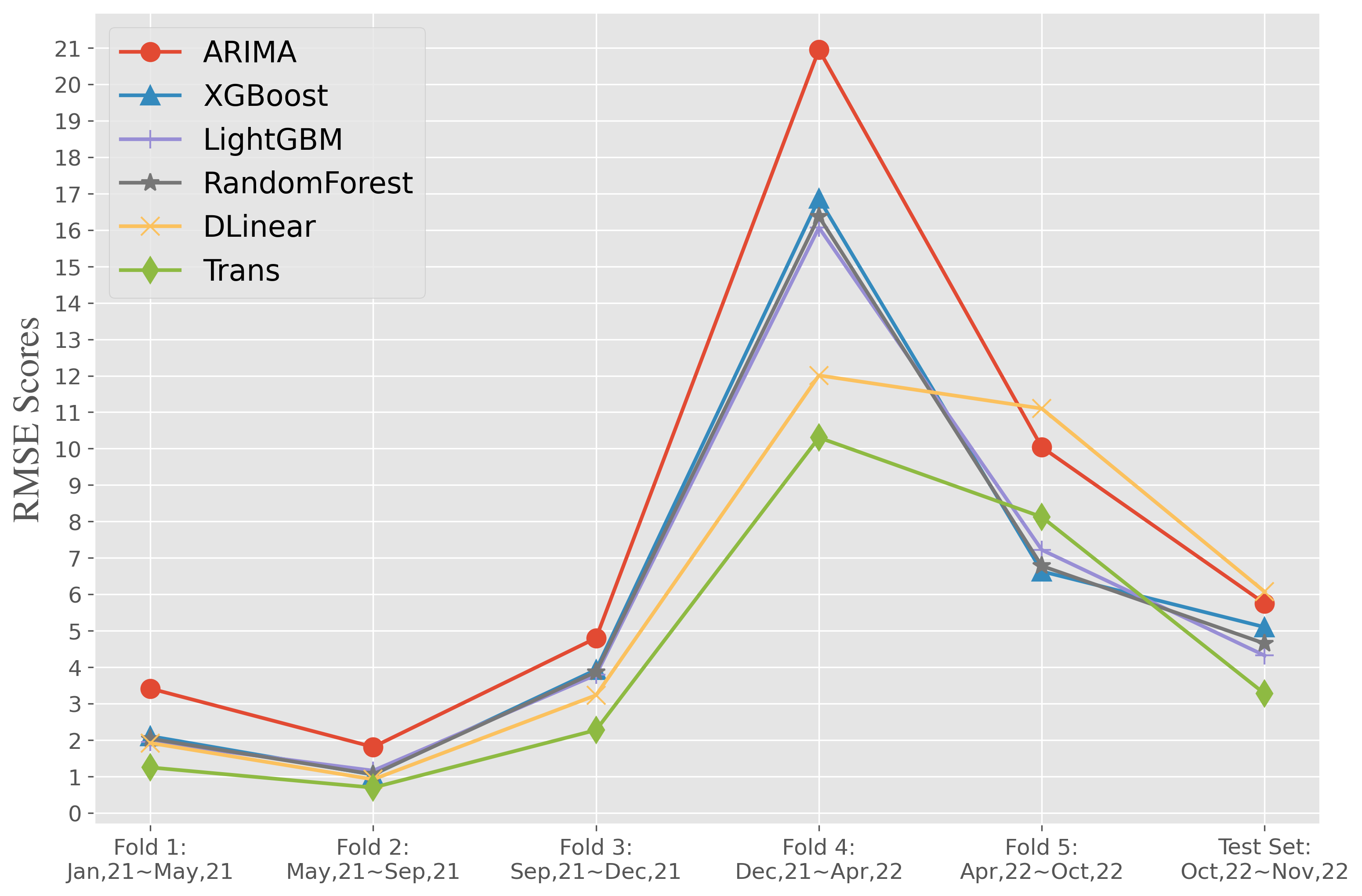}
         \caption{EU RMSE score using I}
         \label{fig:eu_rmse_allfolds_nosupp}
     \end{subfigure}
     \hfill
     \begin{subfigure}[b]{0.45\textwidth}
         \centering
         \includegraphics[width=\textwidth]{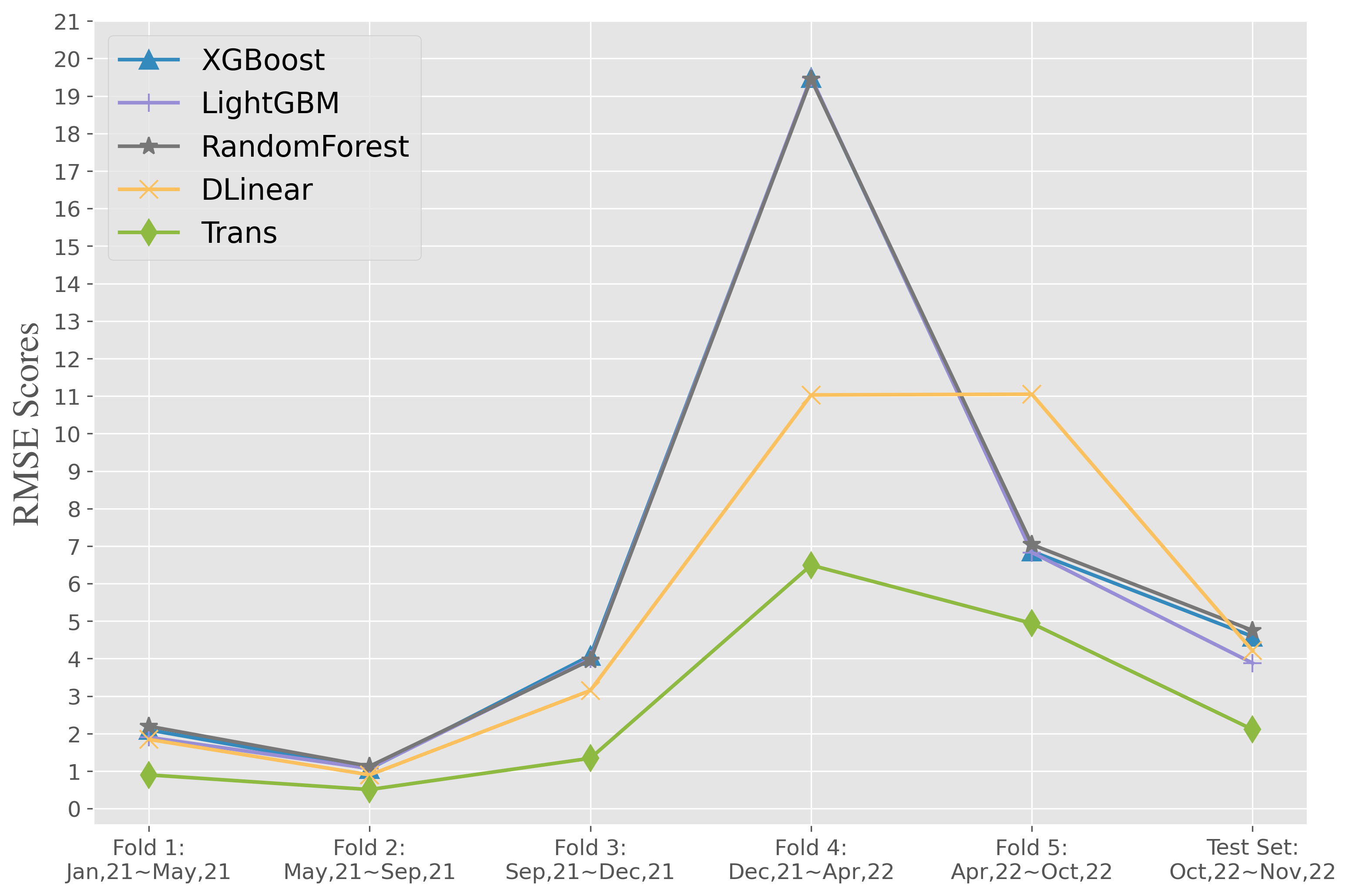}
         \caption{EU RMSE score using IMH}
         \label{fig:eu_rmse_allfolds_supp}
     \end{subfigure}
     \hfill
     \begin{subfigure}[b]{0.45\textwidth}
         \centering
         \includegraphics[width=\textwidth]{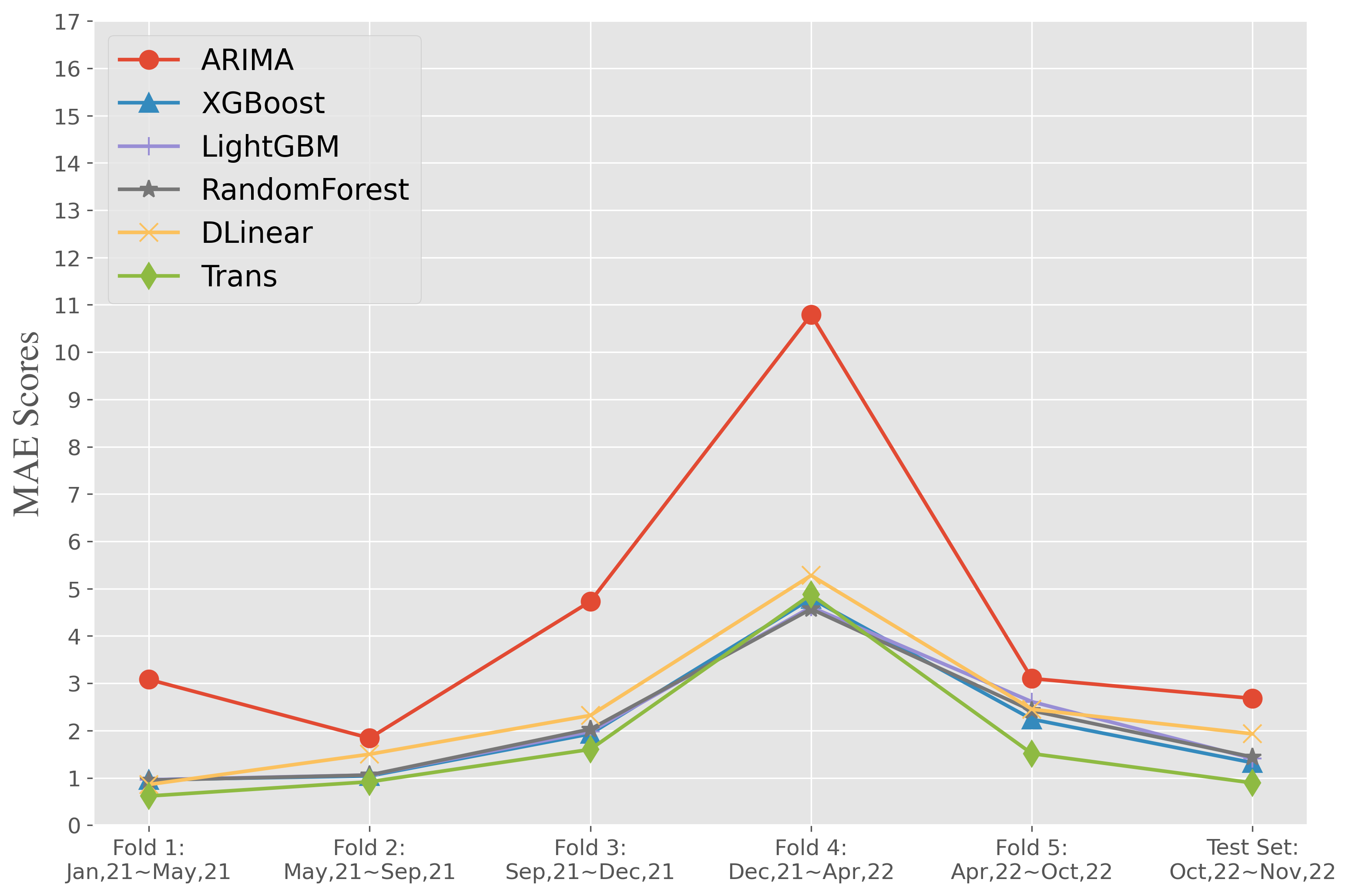}
         \caption{US MAE scores using I}
         \label{fig:us_mae_allfolds_nosupp}
     \end{subfigure}
     \hfill
     \begin{subfigure}[b]{0.45\textwidth}
         \centering
         \includegraphics[width=\textwidth]{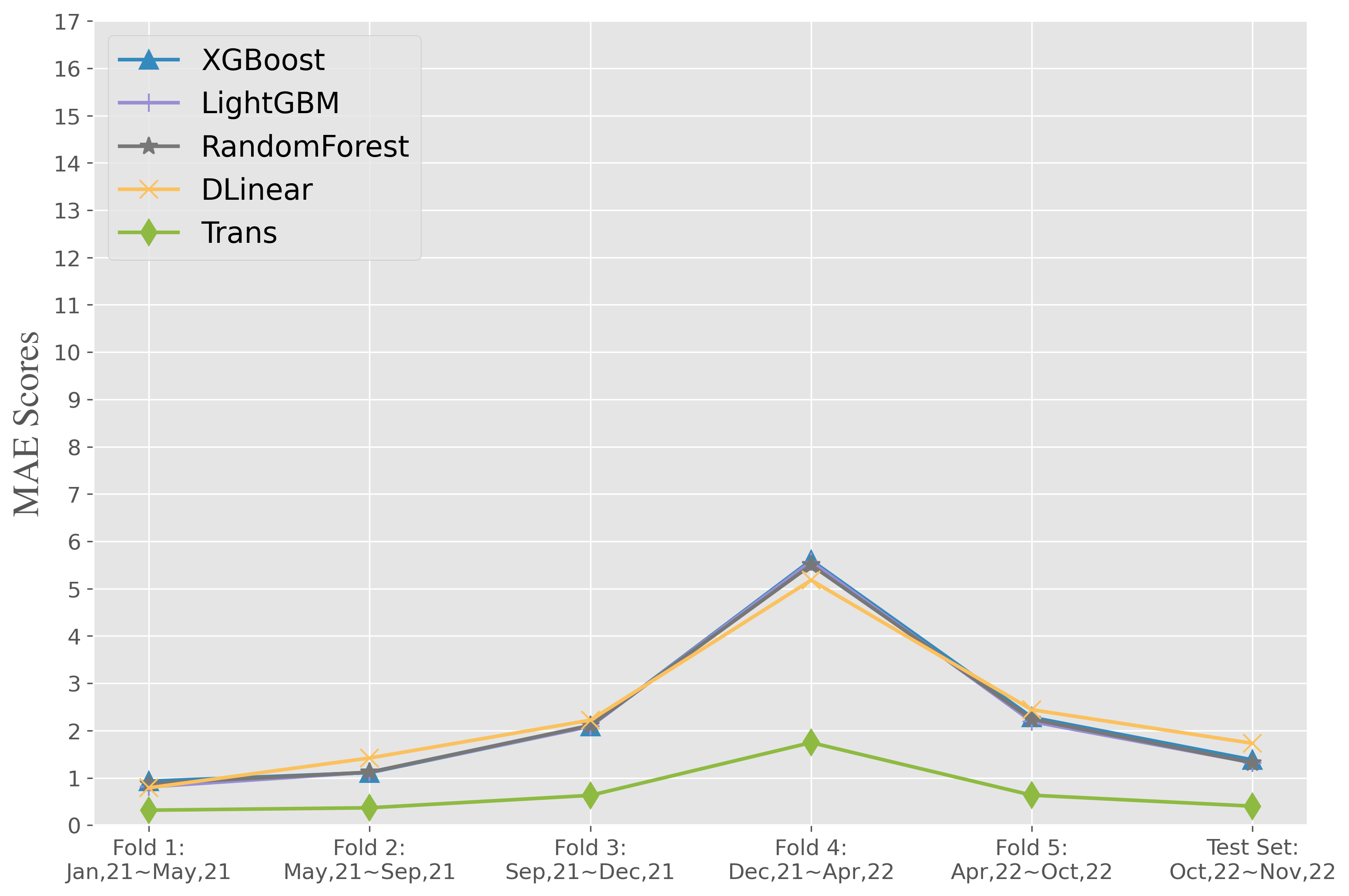}
         \caption{US MAE scores of using IMH}
         \label{fig:us_mae_allfolds_supp}
     \end{subfigure}
     \hfill
     \begin{subfigure}[b]{0.45\textwidth}
         \centering
         \includegraphics[width=\textwidth]{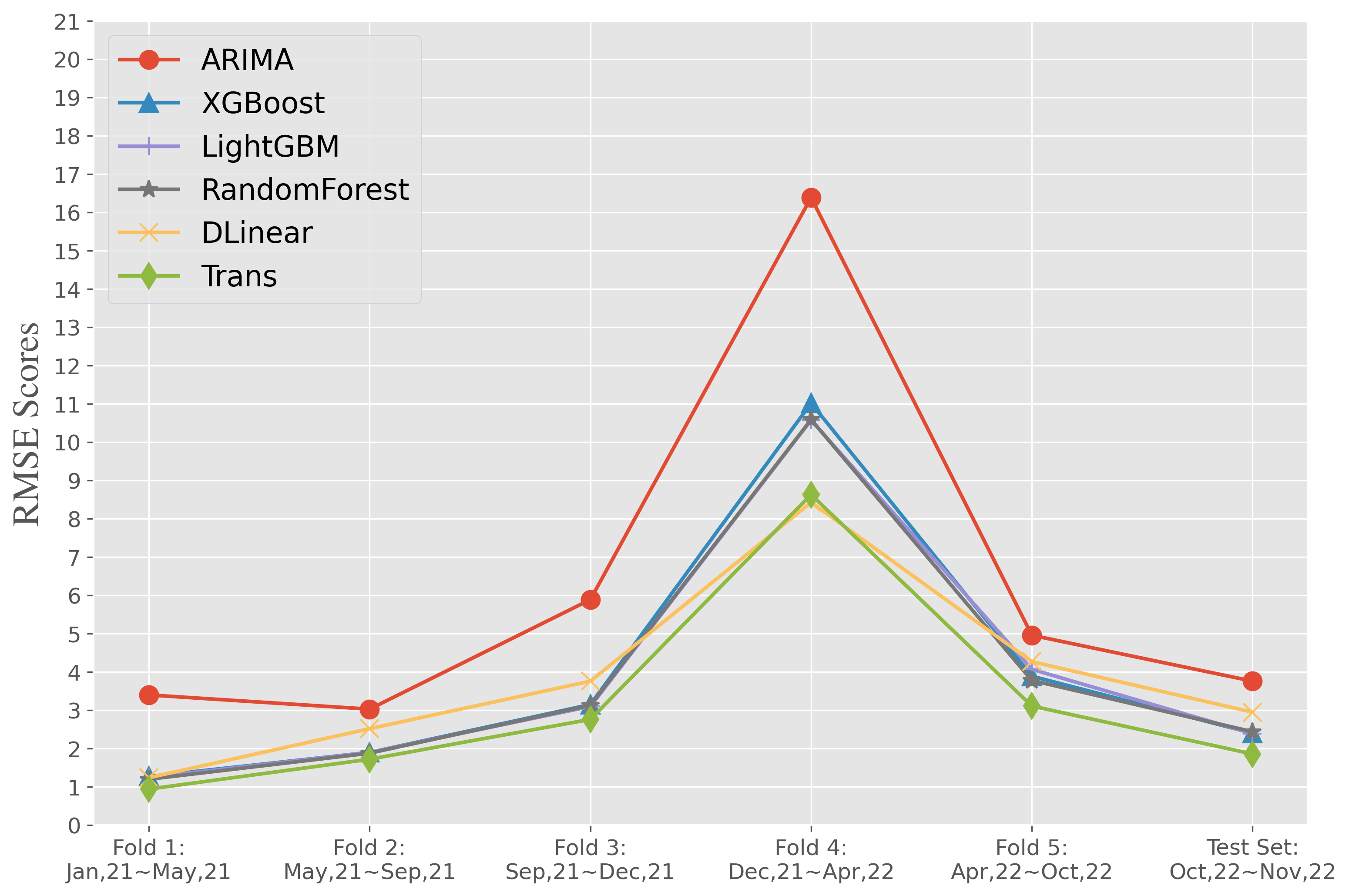}
         \caption{US RMSE scores using I}
         \label{fig:us_rmse_allfolds_nosupp}
     \end{subfigure}
     \hfill
     \begin{subfigure}[b]{0.45\textwidth}
         \centering
         \includegraphics[width=\textwidth]{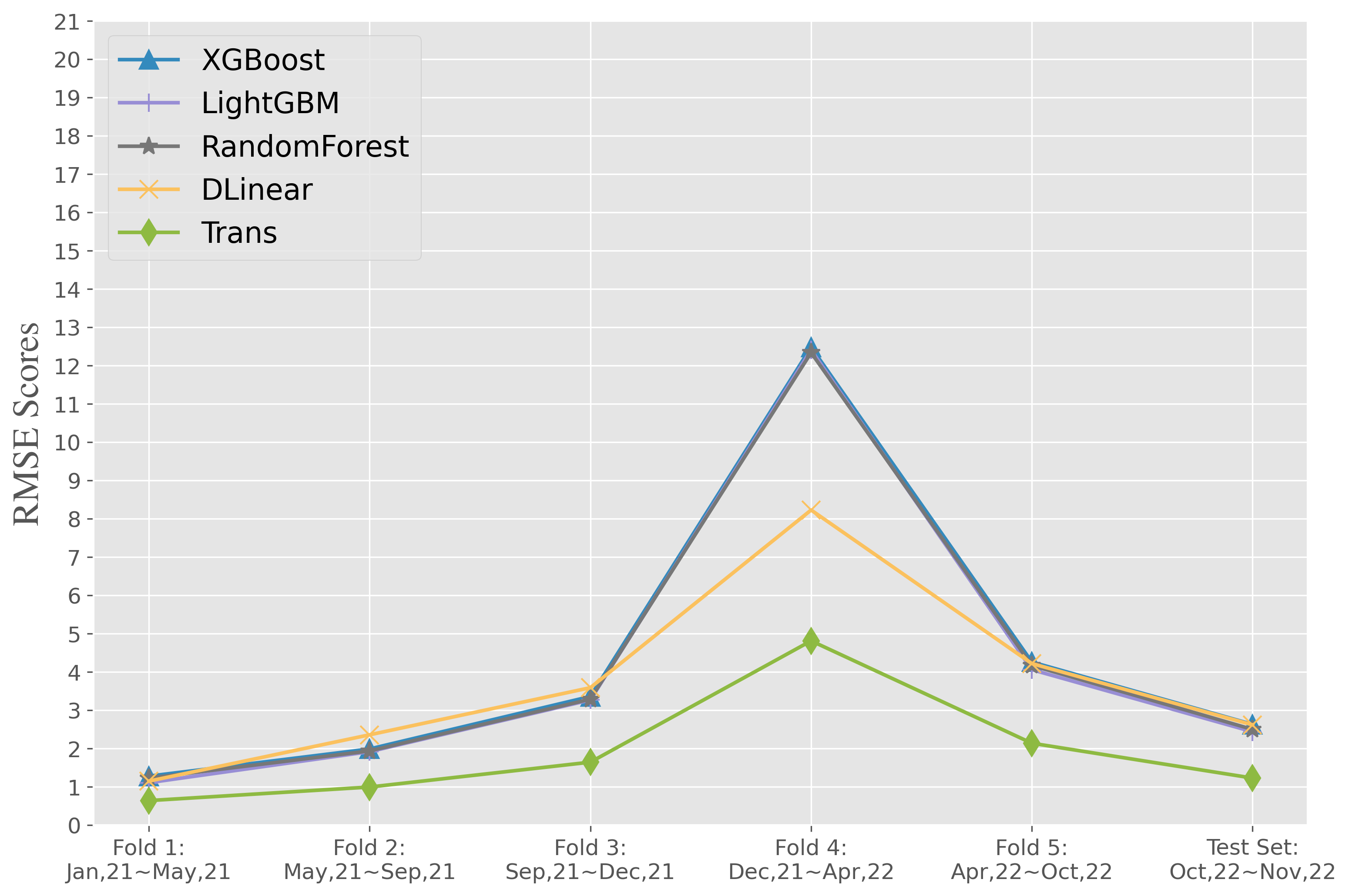}
         \caption{US RMSE scores using IMH}
         \label{fig:us_rmse_allfolds_supp}
     \end{subfigure}
        \caption{\textbf{MAE and RMSE scores of incidence forecasting for EU and US, including ARIMA, Tree-based models, DLinear, and Transformer.} Results on output length of $12$ are reported.}
        \label{fig:all_folds_all_model}
\end{figure}

In Fig.~\ref{fig:all_folds_all_model}, we demonstrate the performance comparison between benchmark models and our backbone model, the vanilla transformer. Our earlier findings suggest that incorporating mobility data and adding the GCN module have a limited impact on forecasting performance. Therefore, we aim to demonstrate the superiority of the transformer over benchmark models to justify our choice of backbone.

To evaluate the resilience of our backbone model to varying data volumes, we adopt a 5-fold cross-validation strategy, with the first fold containing the least data. In Fig.~\ref{fig:all_folds_all_model}, we showcase the model's performance across all 5 folds and the final test set. Deep learning models typically thrive on ample training data, but the unpredictable nature of health policies and resource allocation during the early pandemic stages presents challenges. Thus, we emphasize the model's adaptability to limited data settings, highlighting its potential during initial pandemic phases. Our results demonstrate the model's resilience to data scarcity for COVID-19 incidence forecasting, with MAE and RMSE scores for fold 1 comparable to those of the test set in both regions. However, forecasting case surges, often signaling the emergence of new variants, remains a challenge.

During the fold $4$ test set, the occurrence of the Omicron surge in both EU \cite{ECDCOmicronReport} and US regions \cite{OmicronSurge} results in unexpected incidence patterns. Historical data provide limited insight due to the emergence of the Omicron strain (BA.1) in Botswana and South Africa. Consequently, all models exhibit deteriorated forecasting for the emergent Omicron outbreaks, as depicted in Fig.~\ref{fig:all_folds_all_model}.

Comparing models utilizing only incidence (I) data, as depicted in Fig.~\ref{fig:eu_mae_allfolds_nosupp} and Fig.~\ref{fig:us_mae_allfolds_nosupp}, with models incorporating mortality, hospitalization, and incidence (IMH) data, as shown in Fig.~\ref{fig:eu_mae_allfolds_supp} and Fig.~\ref{fig:us_mae_allfolds_supp}, it reveals that our backbone model, employing MTL, effectively harnesses supplementary variables to enhance forecasting accuracy, particularly evident in fold 4 for both regions. In contrast, the DLinear deep learning model, which disregards spatial connections among sites, does not benefit from MTL. The ARIMA model, employing a univariate forecasting approach, achieves the poorest performance for both regions, owing to its linearity assumption. For shallow machine learning algorithms, boosting and bagging ensemble methods demonstrate similar performance in both regions, as shown in Fig.~\ref{fig:eu_rmse_allfolds_nosupp} and Fig.~\ref{fig:us_rmse_allfolds_nosupp}. By substituting the input from neighboring incidence rates with supplementary variables, we confirm in Fig.~\ref{fig:eu_mae_allfolds_supp}, \ref{fig:eu_rmse_allfolds_supp},~\ref{fig:us_mae_allfolds_supp}
and~\ref{fig:us_rmse_allfolds_supp} that these supplementary variables can be effectively employed for independent incidence prediction. Thus, the application of MTL to forecast supplementary variables represents a closely related task aimed at improving incidence forecasting accuracy.

\section*{Discussion}
The diffusive nature of infectious diseases, as seen in the COVID-19 pandemic, presents a challenge in accurately representing human movement for deep neural networks. Our assessment of Google mobility data's effectiveness in forecasting revealed insufficient evidence to suggest its significant contribution to accuracy. Given limited access to such data, we explored the use of MTL with related clinical data, such as mortality and hospitalization rates, to extract mobility information and enhance predictive performance. Additionally, we propose leveraging the transformer as the backbone model to evaluate the GCN module's efficacy in capturing cross-state and national transmission dynamics, thereby enhancing mobility representation in forecasting tasks. 

In our pilot study, we evaluate the impact of Google Mobility data by integrating it with incidence data in various model structures: Trans, Trans+GCN, Trans+Adp, and Trans+GCN+Adp. Our experiments, comparing IB to I datasets across both regions, reveal insufficient evidence to support the consistent improvement of forecasting performance through mobility data. However, when comparing IMH to I datasets for both regions, we observe that mobility information can indeed be effectively extracted from related clinical data, thereby enhancing forecasting accuracy.

This discrepancy might be due to the nature of Google's mobility data, which is collected via social media and may offer limited insight into actual mobility patterns. In contrast, clinical data reported by hospitals and governments tend to be more reliable, with fewer missing values and errors. Temporally, past hospitalization rates and mortality serve as informative predictors for future incidence trends. An increase in hospitalization rates and mortality within specific time windows signals an expected rise in incidence.

Spatially, a surge in hospitalizations in one sub-region can overflow into neighboring areas, potentially elevating their incidence rates. By leveraging this inter-correlation, the network trained with the MTL strategy is more likely to retrieve mobility patterns from supplementary clinical data. The comparison between IMHB and IMH data further supports that clinical data alone can adequately represent mobility without relying on explicit mobility data.

Building on the insights from our pilot study, we investigate the impact of the GCN module on the transformer backbone across varying data volumes and output lengths. Surprisingly, our findings indicate that the widely used GCN module does not significantly enhance forecasting performance when integrated with the transformer, regardless of whether using a static map, a dynamic map, or both. Unlike its performance on benchmark datasets, the addition of GCN does not improve or, in some cases, even impairs the performance of our datasets. This discrepancy may arise because our datasets contain fewer variables than benchmark ones, making explicit modeling with the GCN module unnecessary for forecasting tasks.

While minor improvements were observed for the EU dataset with dynamic maps, no significant enhancements were detected for the US dataset. Interestingly, the GCN module exhibited slight performance gains for longer output lengths (24 and 36) and in scenarios of data scarcity (fold 1 and fold 2). This suggests that the GCN module may hold promise in aiding mobility retrieval, particularly in challenging forecasting tasks such as long-term predictions and scenarios with limited data. 

Combining these observations, we suggest that the GCN module is only effective in modeling complex interactions between variables under challenging tasks. However, within our experimental framework, the GCN module fails to demonstrate a significant contribution to forecasting compared to the vanilla transformer, emphasizing the paramount importance of temporal information in the forecasting process.

Though the GCN module offers limited improvement in forecasting accuracy, the dynamic maps generated from the spatial attention module can offer insight into mobility trends. Comparing the dynamic maps with the static ones, we find that the static maps are more connected and heavily weighted than the dynamic ones, suggesting that the addition of dynamic maps in Trans+GCN+Adp may have little impact on the model's predicting power, giving a likely explanation of the deteriorated performance of Trans+GCN+Adp, compared to the Trans+Adp model. Assessing the mobility indicator across folds and output length, we observe that the connectivity and average weights vary across folds and output lengths for both EU and US, and the highest connectivity and weights are more likely to be found in longer output lengths and earlier fold, showing that the dynamic maps can be effective for more challenging forecasting tasks.

We observe that the mobility indicator of Trans+Adp models can represent the change in the human movement for lockdown orders concurrently issued by more than one country. We believe that the recurring lockdown orders may trigger a loss in compliance due to the pandemic fatigue. We have tested different time spans for the pre- and post-lockdown periods to check if mobility indicator fluctuation exists. As a result, we observe that the lockdown duration does not contribute much to the change in mobility indicator, and $24$ days seems to adequately capture the change.

In terms of forecasting performance, we observe that all of our models, including the vanilla transformer, Trans+GCN, Trans+Adp, and Trans+GCN+Adp, can all outperform the benchmark models, DLinears and XGBoost. Moreover, our models faithfully retrieve the rebound in incidence after lockdown orders for both the EU and the US, showing the importance of spatial connection modeling, explicitly or implicitly. 

In interpreting this study's findings, it is important to consider the limitations inherent in our approach. Due to limited access to mobility data on the regional scale, we limit the comparison between mobility data and the IMH dataset in the pilot study and exclude mobility data from the main experiments. Moreover, mobility data from other sources may be more informative than Google mobility data, which is not assessed in the current study. However, we have confirmed that Google Mobility has been adopted in many studies \cite{wardForecastingSARSCoV2Transmission2022, liuAssociationsChangesPopulation2021} as a proxy for human movement, and thus, we entrust this data source to provide enough support to our findings.

Additionally, the exclusion of sites in the US regions due to excessive missing data introduces potential bias in the results, such as Alaska, Hawaii, American Samoa, Guam, the Northern Mariana Islands, Puerto Rico, and the U.S. Virgin Islands from the US dataset. As a result, we have limited our focus in the continental states of the US. In the EU region, there is a significant amount of missing data for hospitalization entries due to the fact that, for certain countries, hospitalization data are reported on a weekly basis. The imputation for hospitalization may cause shifts in the data distribution and cause deterioration in performance in our model.

It is also worth noting that our model requires a sufficiently large training set, such as the fold $1$ training set comprising approximately $300$ samples. A target dataset containing extremely scarce data for training, such as only the first month of confirmed cases for all sites, will make the model incapable of accurate forecasting.

For future investigations, we recommend conducting a comparative analysis between incorporating mobility data from other sources and leveraging regional relationships to better understand disease spread. Furthermore, the development of imputation methods specifically tailored to handle case recording of infectious diseases is crucial, as real-world datasets often require data cleaning, and discarding flawed data may not be the optimal solution. Lastly, we propose the use of Transfer Learning (TL) as a means to forecast under conditions of data scarcity, enabling reliable predictions at the early stages of a pandemic.


\section*{Methods}
\subsection*{Transformer-Based GCN-Assisted COVID-19 Forecasting Model}
In this study, we present a comprehensive overview of the mathematical symbols used, which can be found in Supplementary Table 6. The proposed framework, as depicted in Fig.~\ref{fig:network}, comprises three key modules: 1) a transformer backbone for processing temporal information; 2) a dynamic learning module integrated into the transformer backbone to generate adjacency matrices representing spatial connections; and 3) a global graph message module that combines temporal information with spatial associations.

\subsubsection*{Transformer backbone}
To capture the temporal association, we use the encoder-decoder structure of the transformer model where the input $\mathbf{X}$ is embedded with fully-connected layers to generate features with arbitrary dimension $D$. To account for the positional information within the sequence, we employ positional encoding $\mathbf{P} \in \mathbb{R}^{N \times T \times D}$, as described in Eq. \eqref{eq:PE}, where $i$ and $j$ denote the position in the sequence and dimension, respectively. The addition of positional encoding and embeddings results in the final input, denoted as $\mathbf{Z}^{(0)} \in \mathbb{R}^{N \times T \times D}$.
\begin{align}
    \mathbf{P}_{(i, 2j)} = \sin \left(\frac{i}{10000^{2j/D}}\right), \mathbf{P}_{(i, 2j+1)} = \cos \left(\frac{i}{10000^{2j/D}} \right). \label{eq:PE}
\end{align}

The core structure of the encoder includes three key components: 1) an attention module that facilitates associations between non-adjacent positions in the sequence; 2) a feed-forward module consisting of two linear transformations with an in-between ReLU activation, applied individually and uniformly to each position; and 3) residual connections that enable the addition of the input features to the normalized output, promoting gradient flow. These components form an encoder stack, and the number of stacks needed in the encoder is determined by the complexity and volume of the data. Since the feed-forward module and residual connections are straightforward, we provide a detailed exposition of the attention mechanism, as shown in Eq. \eqref{eq:temporal_embed} and \eqref{eq:attention}. 

We begin by embedding the input to obtain the query $\mathbf{Q} \in \mathbb{R}^{N \times T \times D_k}$, key $\mathbf{K} \in \mathbb{R}^{N \times T \times D_k}$, and value $\mathbf{V} \in \mathbb{R}^{N \times T \times D_v}$, using linear layers with learnable parameters $\mathbf{W}_q, \mathbf{W}_k \in \mathbb{R}^{D \times D_k}$ and $\mathbf{W}_v \in \mathbb{R}^{D \times D_v}$. The dimensions are adjusted to $D_k$ and $D_v$ to accommodate the multiple heads employed in the attention mechanism. For the sake of simplicity, we omit the head dimension in the equations to maintain clarity in the notation, i.e.,
\begin{align}
     \mathbf{Q} &= \mathbf{W}_q\mathbf{Z}^{(l)}, \quad \mathbf{K} = \mathbf{W}_k\mathbf{Z}^{(l)}, \quad  \mathbf{V} = \mathbf{W}_v\mathbf{Z}^{(l)}. \label{eq:temporal_embed}
\end{align}
We compute the attention matrix by taking the dot-product between the query and key along the temporal dimension. This matrix is subsequently normalized using the softmax function $\sigma(\cdot)$ to obtain the attention score. By multiplying the score with the value vector, we integrate temporal information from adjacent and non-adjacent positions.
\begin{align}
    \mathbf{Z}_t^{(l)} = \sigma\left(\frac{\mathbf{Q}\mathbf{K}^T}{\sqrt{D_k}}\right)\mathbf{V}. \label{eq:attention}
\end{align}

The decoder of the transformer consists of a linear layer with learnable parameter $\mathbf{W}_o \in \mathbb{R}^{T \times F}$, which projects the feature obtained from the last encoder stack $L$ onto the desired time steps $F$. Subsequently, this feature is passed through a generator linear layer, $\mathbf{W}_g \in \mathbb{R}^{D \times D_o}$, to transform it to the desired dimension $D_o$.
\begin{align}
    \hat{\mathbf{Y}} = \mathbf{W}_g(\mathbf{W}_o\mathbf{Z}^{(L)}).
\end{align}

\subsubsection*{Dynamic Spatial Connection Learning}
For each region of interest, two types of spatial connections are considered. First, we account for the physical distance between locations, which is inversely related to the likelihood of travel. Second, we incorporate hidden correlations beyond distance by considering the influence of the main or supplementary variables from neighboring locations. To establish the physical-distance-based connection, we generate an adjacency matrix $A_g$ using a thresholded Gaussian kernel on pairwise distances between locations \cite{liDiffusionConvolutionalRecurrent2018}. The element in the row $i$ and column $j$ of this matrix denotes the edge weight $W_{ij}$ between node $i$ and $j$ acquired using: 
\begin{align}
    W_{ij} = 
    \begin{cases}
         \exp{(-\frac{\text{dist}(v_i, v_j)^2}{\sigma^2})}, &\text{if dist}(v_i, v_j) \le \kappa, \\
         0, &\text{otherwise},
    \end{cases} \label{eq:physical_distance}
\end{align}
where $\sigma$ and $\kappa$ represent the standard deviation of distances and pre-set noise threshold. The longitude and latitude are gathered for locations using Google Map API and the distance between every pair of sites are derived from their coordinates.

We utilize the attention mechanism to capture non-distance-related correlations and estimate spatial associations. To achieve this, we obtain node query, key, and value vectors, denoted as $\mathbf{Q}_n, \mathbf{K}_n$, and $\mathbf{V}_n$ respectively, by applying linear layers to the stack input $\mathbf{Z}^{(l)}$. Subsequently, we apply the attention mechanism to the spatial dimension, resulting in the generation of the node attention score $\mathbf{M}_n \in \mathbb{R}^{N \times N}$ as:
\begin{align}
    \mathbf{M}_n = \sigma\left(\frac{\mathbf{Q}_n\mathbf{K}_n^T}{\sqrt{D_k}}\right). \label{eq:node_att_self} 
\end{align}
The softmax function, shown as:
\begin{align}
    \sigma(\mathbf{s}) = \frac{\exp{(\mathbf{s}_i)}}{\sum_{j=1}^{N} \exp{(\mathbf{s}_j)}}, \quad \text{for} \quad i = 1,..., N, \label{eq:softmax}
\end{align}
is used to normalize attention, which ensures that the probability for each row vector $\mathbf{s}$ is constrained within the range $(0, 1)$. Because row entries for the score can never reach one or zero, $\mathbf{M}_n$ is a dense matrix filled with trivial weights. To address this redundancy, we introduce a hard thresholding function to sparsify $\mathbf{M}_n$ and obtain the output matrix $\mathbf{A}_s^{(l)}$, i.e,
\begin{align}
    \mathbf{A}_s^{(l)} = g(\mathbf{M}_n, \tau) =
    \begin{cases}
        0, &\text{if} \quad \mathbf{M}_{n, (i, j)} < \tau, \\
        \mathbf{M}_{n, (i, j)},&\text{if} \quad \mathbf{M}_{n, (i, j)} \geq \tau. \label{eq:thresholding}
    \end{cases}
\end{align}
We define $\mathbf{M}_{n, (i, j)}$ as the element in the $i$-th row and $j$-th column of the matrix, and $\tau$ represents the pre-defined threshold. The sparsification of the generated adjacency matrix is governed by a pre-set sparsity ratio $\rho$, as outlined in Algorithm \ref{alg:update_sparsity}

\begin{algorithm}
\caption{Sparsification of the generated adjacency matrix}\label{alg:beta_update}
\textbf{Input}: $\mathbf{M}_n$: nodes attention score \\
\textbf{Output}:  $\mathbf{A}_s^{(l)}$:  generated adjacency matrix 
\begin{algorithmic}[1]
\State $\tau \gets 1/N$  \Comment{if the connection weight is less than uniformly selected, remove it}
\If{$\mathbf{A}_g$ exists}
    \State $\rho \gets (\text{No. of non-zero connections in } \mathbf{M}_n) / (N \times N) $
\Else
    \State $\rho \gets 0.001 $ \Comment{if geographical adj. matrix exists, its sparsity is used; otherwise $\rho$ is a hyper-param set to $0.001$ as initiation}
\EndIf
\\
\For{each node attention score}:
    \State $\mathbf{A}_s^{(l)} \gets g(\mathbf{M}_n, \tau)$
    \State $r = (\text{No. of non-zero connections in } \mathbf{A}_s^{(l)}) / (N \times N)$
    \If{$r > \rho$}
        \State $\mathbf{A}_s^{(l)} \gets \mathbf{A}_s^{(l)}$ \Comment{if not met the pre-set sparsity ratio, use hard-threasholding}
    \Else
        \State $\mathbf{A}_s^{(l)} \gets \mathbf{M}_n$
    \EndIf
\EndFor
\end{algorithmic}
\label{alg:update_sparsity}
\end{algorithm}

\subsubsection*{Graph Message Passing – Propagation}
In this study, we employ graph neural networks to capture the spread of epidemic diseases at the population level. These networks iteratively update node features to model the influence of population-level mobility on disease transmission across locations. The propagation process in graph neural networks revolves around convolution layers, which consist of two stages: message passing and aggregation. For our investigation, we adopt the Graph Convolutional Networks (GCN) as a classical GNN layer. GCN utilizes linear layers and normalization for connected nodes in the message passing stage, followed by a simple summation in the aggregation stage. By considering the geographical adjacency matrix and denoting the linear layer weight as $\Theta$, the propagation can be expressed as follows:
\begin{align}
    \tilde{\mathbf{A}}_g &= \mathbf{D}^{-\frac{1}{2}}\mathbf{A}_g\mathbf{D}^{-\frac{1}{2}}, \nonumber \\
    \mathbf{h}_g^{(l)} &= \tilde{\mathbf{A}}^g\mathbf{Z}_t^{(l)}\Theta. 
\end{align}
To simulate the ripple effect of virus transmission and capture broader neighborhood interactions, we extend the propagation process. For example, encountering an individual who recently had contact with a COVID-19 patient significantly increases the likelihood of infection. By leveraging the capabilities of GCN, we can propagate information not only from directly connected neighbors but also from neighbors $K$ hops away. This is achieved through the nesting of convolutional layers, as demonstrated below:
\begin{align}
    \mathbf{h}_g^{(l)} = \sum_{k=0}^{K} {\tilde{\mathbf{A}}_g}^k \mathbf{Z}_t^{(l)} \Theta_k.
\end{align}
To leverage both geographical and generated adjacency matrices, we incorporate nested GCNs with a concatenation module, denoted as $\textbf{CONCAT}(\cdot)$, to aggregate features from two adjacency matrices. This integration is depicted as follows:
\begin{align}
    \mathbf{h}^{(l)} &= \mathbf{W}_x \cdot \textbf{CONCAT}\left( \sum_{k_g=0}^{K} {(\tilde{\mathbf{A}}_g)}^{k_g} \mathbf{Z}_t^{(l)} \Theta_{k_g}, \quad \sum_{k_s=0}^{K}{(\tilde{\mathbf{A}}_s^{(l)})}^{k_s} \mathbf{Z}_t^{(l)} \Theta_{k_s} \right).    
\end{align}
The feature output from GCNs is subsequently passed through the ReLU activation function to obtain the updated feature for the current block, as depicted below:
\begin{align}
    \mathbf{Z}^{(l+1)} \gets \text{ReLU}(\mathbf{h}^{(l)}) + \mathbf{Z}_t^{(l)}.
\end{align}

\subsubsection*{Data Pre-processing}
To select the optimal input and output lengths for experiments, we establish two groups representing the seasonal and the non-seasonal forecasting scenarios. In the non-seasonal group, the input length is set to $12$, while the output lengths of $3$, $6$, $12$, $24$, and $36$ represent short-term to long-term forecasting. Given the daily frequency of the data, the non-seasonal scenario assumes the absence of a weekly pattern. Conversely, the seasonal group has an input length of $14$ and output lengths of $2$, $7$, and $14$, considering the influence of a weekly pattern on our Deep learning model's predictions. Consequently, the total timesteps of the test set vary for each input and output length pair. To ensure consistent output timesteps across all experiments, we have provided a comprehensive overview of the number of samples and expected timesteps for each forecasting window in Supplementary Table 7.

\subsubsection*{Hyper-parameter Tuning}
Achieving valid deep learning models heavily relies on hyper-parameter tuning, a crucial step in our network development. To identify the most robust model, we utilize the validation set. The transformer structure employed in our approach necessitates a learning rate warmup to optimize results, with the maximum learning rate and the number of training steps to reach it being key tuning parameters. Through extensive experimentation, we have determined that a peak learning rate of $0.001$ and $20000$ training steps yield the best forecasting performance. Additionally, we have observed that our model exhibits reasonable robustness to changes in the warmup process. Researchers working with different datasets are advised to initially adopt our hyper-parameter settings.

In addition to the learning rate, the manipulation of the adjacency matrix is governed by hyper-parameters. The ``set\_diag'' parameter controls whether diagonal weights are enabled for self-looping, while ``undirected'' indicates whether the matrix is symmetric. The ``truncate'' parameter determines whether weights below a certain threshold are set to zero, and ``threshold'' represents the pre-set ratio used in this decision, adjusted according to the number of weights.

Regarding the geographical matrix, we assume it to be non-symmetric, without self-loops, and with no weights truncated, allowing it to preserve information optimally. Conversely, we enforce conditions for the generated matrix to be symmetric, equipped with self-loops, and with truncated weights present. This approach ensures that the matrix remains resilient to noise amplification from the softmax function by discarding insignificant weights. Furthermore, the self-loop property guarantees that the generated matrix is at least as dense as the identity matrix, enhancing its overall effectiveness.

\subsection*{Sensitivity Analysis}
\subsubsection*{Shallow Machine Learning}
In our study, we employ shallow learning algorithms, namely XGBoost \cite{chenXGBoostScalableTree2016}, LightGBM \cite{keLightGBMHighlyEfficient2017a}, and RandomForest \cite{breimanRandomForests2001}, to train on the incidence rate of COVID-19. As these algorithms are not conventional time series forecasting models, we deviate from the sliding-window technique explained above. 
To identify the optimal model, we employ a grid search algorithm that explores three key parameters: ``max\_depth,'' ``n\_estimators,'' and ``learning\_rate,'' across each fold and the entire training set. For RandomForest, the ``max\_features'' parameter is searched and the ``learning\_rate'' is omitted. This process entails training and validating a total of $27$ parameter configurations for each output length within every fold as shown in the Supplementary Table 8. Subsequently, the best models from each fold, along with the test set, are evaluated using holdout testing samples.

The organization of data poses a challenge due to the dimensional constraints imposed on input data, which necessitates a two-dimensional table format. Consequently, when dealing with multi-site data that includes supplementary variables for each site, it becomes impractical to fit such data into shallow-learning algorithms. To address this issue, we adopt a two-step approach to evaluate the predictive power of neighboring locations and supplementary variables separately. First, we utilize historical incidence rates from $N-1$ neighboring sites to predict the incidence rates at the selected location. Secondly, we aggregate historical values of all supplementary variables to forecast the incidence rates specifically for the chosen location. The input data format is depicted in Supplementary Fig.5.

\subsubsection*{ARIMA and Linear Layer Deep Learning}
For preliminary univariate analysis, we employed an autoregressive integrated moving average (ARIMA) model, fitted using a modified Hyndman-Khandakar algorithm. The ``AutoArima" function from the ``scikit-learn" package facilitated the search for crucial parameters, including the optimal order of the auto-regressive (AR) model ($p$) and the order of the moving average (MA) model ($q$). To fit the ARIMA model for the daily incidence rate at each location, we employed K-fold progressive cross-validation as previously described. Metrics were computed on the test set, pooled, and averaged over sites to obtain the final result. In comparison to the Deep Learning model, we set the maximum values for $p$ and $q$ equal to the input length, which was $14$ for seasonal experiments and $12$ for non-seasonal experiments. 

In addition to the ARIMA model, we incorporated linear layers to assess the effectiveness of our approach. Recent research has demonstrated that simple linear layers can achieve comparable performance to complex model designs for multivariate time series forecasting \cite{zengAreTransformersEffective2023}. However, the role of the linear layer is to map input features to the output space, $\mathbb{R}^{a \times b \times \cdots \times \text{input\_dim}} \to \mathbb{R}^{a \times b \times \cdots \times \text{output\_dim}}$, disregarding dimensions other than the last. In contrast, our proposed method explicitly models spatial and temporal dimensions separately and implicitly accounts for supplementary variables if present. To accommodate this, we assign a linear layer to each site, and aggregate results for each variable in the presence of supplementary variables. The data preprocessing steps align with our proposed method, and we select the DLinear structure based on its superior performance on our dataset.

\subsection*{Data Availability} \label{subsec:data_avail}
The data on cases, deaths, and hospitalizations for both regions are provided by the Johns Hopkins University (JHU) data tracking program \cite{dongInteractiveWebbasedDashboard2020}. 
Cases and deaths can be accessed at the GitHub repository \url{https://github.com/CSSEGISandData/COVID-19} and the hospitalizations at U.S. Department of Health and Human Services with \url{https://healthdata.gov/Hospital/COVID-19-Reported-Patient-Impact-and-Hospital-Capa/g62h-syeh}. The hospitalization for the EU merged JHU data with data from Our Wolrd in Data (OWID) at \url{https://github.com/owid/covid-19-data} to reduce missing values. 
The Latitude and longitude of sites are obtained using Google Map API at \url{https://www.gps-coordinates.net/}. 
The COVID-19 response data is obtained at the European Centre for Disease Prevention and Control (ECDC) with \url{https://www.ecdc.europa.eu/en/publications-data/download-data-response-measures-covid-19} for the EU and Centres for Disease Control and Prevention (CDC) for US with \url{https://data.cdc.gov/Policy-Surveillance/U-S-State-and-Territorial-Stay-At-Home-Orders-Marc/y2iy-8irm}. 
The shallow machine learning algorithms are implemented using package xgboost (\url{https://xgboost.readthedocs.io/en/stable/python/python_intro.html}), lightgbm (\url{https://lightgbm.readthedocs.io/en/stable/Python-API.html}) and sklearn (\url{https://scikit-learn.org/stable/modules/generated/sklearn.ensemble.RandomForestRegressor.html}) for XGBoost, LightGBM and RandomForest. 
The ARIMA algorithms are implemented using package sktime (\url{https://sktime-backup.readthedocs.io/en/v0.13.1/api_reference/auto_generated/sktime.forecasting.arima.AutoARIMA.html}).

$28$ countries included in the EU region are Austria, Belgium, Bulgaria, Croatia, Cyprus, Czechia, Denmark, Estonia, Finland, France, Germany, Greece, Hungary, Ireland, Italy, Latvia, Lithuania, Luxembourg, Malta, Netherlands, Norway, Poland, Portugal, Romania, Slovakia, Slovenia, Spain and Sweden. 

$49$ states included in the US region are Alabama, Arizona, Arkansas, California, Colorado, Connecticut, Delaware, District of Columbia, Florida, Georgia, Idaho, Illinois, Indiana, Iowa, Kansas, Kentucky, Louisiana, Maine, Maryland, Massachusetts, Michigan, Minnesota, Mississippi, Missouri, Montana, Nebraska, Nevada, New Hampshire, New Jersey, New Mexico, New York, North Carolina, North Dakota, Ohio, Oklahoma, Oregon, Pennsylvania, Rhode Island, South Carolina, South Dakota, Tennessee, Texas, Utah, Vermont, Virginia, Washington, West Virginia, Wisconsin, Wyoming.

\subsection*{Code Availability}
Experiments are conducted on a machine with eight NVIDIA GeForce RTX 2080 GPUs. Code is implemented with Ubuntu 18.04, PyTorch 1.10.1, and Python 3.8.12. Code with pre-processed data are available at \url{https://github.com/SuhanG17/MTL-Cov-GCN.git}.

\bibliography{COVID19_SR}

\section*{Acknowledgements}
This work is supported in part by the National Natural Science Foundation of China under Grant Nos. 62276127.

\section*{Author contributions statement}
S.G. developed the method and conducted the experiments in Sensitivity Analysis. Z.X. conducted the experiments in the Pilot Study and Main Experiments. F.S. and J.Z. co-led the overall project. All authors (S.G., Z.X., F.S., and J.Z.) participated in paper writing.

\section*{Additional information}
The authors declare no competing interests. This manuscript contains supplementary material available in the supplementary information PDF file.

\end{document}


%
%

\section*{Pilot study: t-test results}
The P-values of the one-sided two-sample independent t-test on MAE and RMSE scores for the pilot study are shown in Supplementary Table \ref{tab:p-value_pilot}.

\begin{table}[h]
\centering
\caption{P-values for one-sided two-sample independent t-test on MAE and RMSE scores, $\alpha$=$0.05$}
\begin{tabular}{@{}lccccc@{} } 
 \toprule
 &  & IB vs. I & IMH vs. IB & IMH vs. I & IBMH vs. IMH \\
 \cmidrule[0.4pt](r{0.125em}){1-1}
 \cmidrule[0.4pt](r{0.125em}){2-2}
 \cmidrule[0.4pt](r{0.125em}){3-3}
 \cmidrule[0.4pt](r{0.125em}){4-4}
 \cmidrule[0.4pt](r{0.125em}){5-5}
 \cmidrule[0.4pt](r{0.125em}){6-6}
 
 \multirow{2}{*}{MAE} & EU & 0.452 & \bf{0.000} & \bf{0.000} & 1.000 \\
                      & US & 1.000 & \bf{0.000} & \bf{0.000} & 1.000 \\

 \cmidrule[0.4pt](r{0.125em}){1-1}
 \cmidrule[0.4pt](r{0.125em}){2-2}
 \cmidrule[0.4pt](r{0.125em}){3-3}
 \cmidrule[0.4pt](r{0.125em}){4-4}
 \cmidrule[0.4pt](r{0.125em}){5-5}
 \cmidrule[0.4pt](r{0.125em}){6-6}
 
 \multirow{2}{*}{RMSE} & EU & \bf{0.000} & \bf{0.000} & \bf{0.000} & \bf{0.000} \\
                       & US & 1.000 & \bf{0.000} & \bf{0.020} & 1.000\\
 
 \bottomrule
\end{tabular}
\label{tab:p-value_pilot}
\end{table}

\section*{Main experiments: Analysis on weekly trend}
By screening the data, we observe that the missing pattern repeats weekly for some subregions: some data entries are missing for the weekend, while for others, data entries are missing for weekdays. To avoid sharp changes in the input data slices, we propose using a length of 12 steps as the input length. To assess the impact of input and output length on model performance, we conduct experiments with different input-output time steps. When the input window size fails to span a multiple of 7, weekly information is compromised. In contrast, preserving weekly information is possible when the input length is set to 14, with output lengths of 2, 7, and 14 steps. As shown in Supplementary Fig. \ref{fig:weekly_trend}, comparing the output performance between data with and without weekly trends, we find no visual difference. The two-sided t-test with $\alpha=0.05$ indicates no statistically significant evidence when comparing the two types of data for all four model structures in both MAE ($P=0.916, 0.663, 0.923, 0.882$) and RMSE ($P=0.991, 0.912, 0.984, 0.998$) scores. We conclude that the periodic pattern offers limited insights into the disease transmission process. Therefore, we decided to use $12$ steps as the input length and $3, 6, 12, 24, 36$ steps as the output length.

\begin{figure}[h]
     \centering
     \begin{subfigure}[b]{0.45\textwidth}
         \centering
         \includegraphics[width=\textwidth]{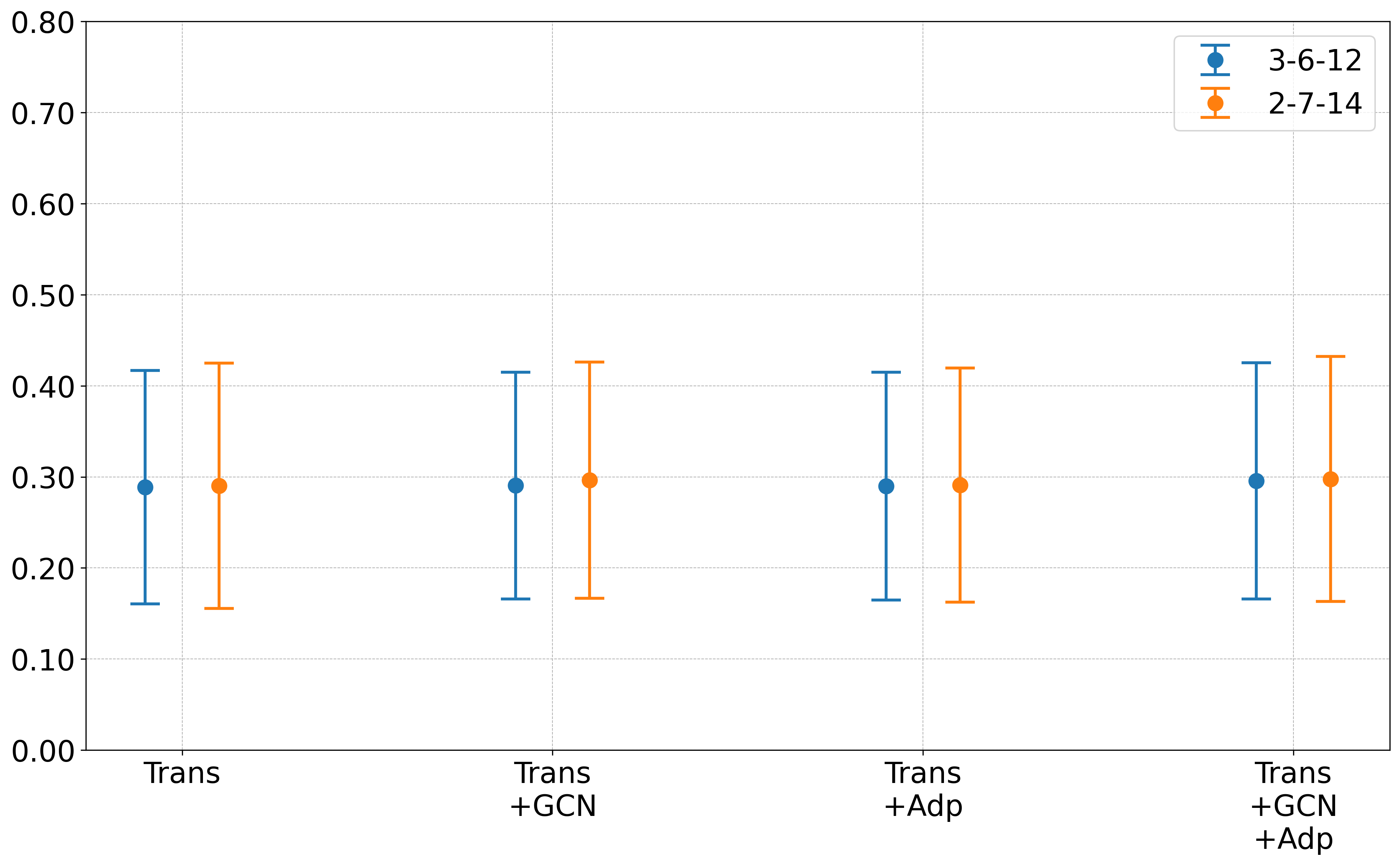}
         \caption{MAE scores}
         \label{fig:weekly_mae}
     \end{subfigure}
     \hfill
     \begin{subfigure}[b]{0.45\textwidth}
         \centering
         \includegraphics[width=\textwidth]{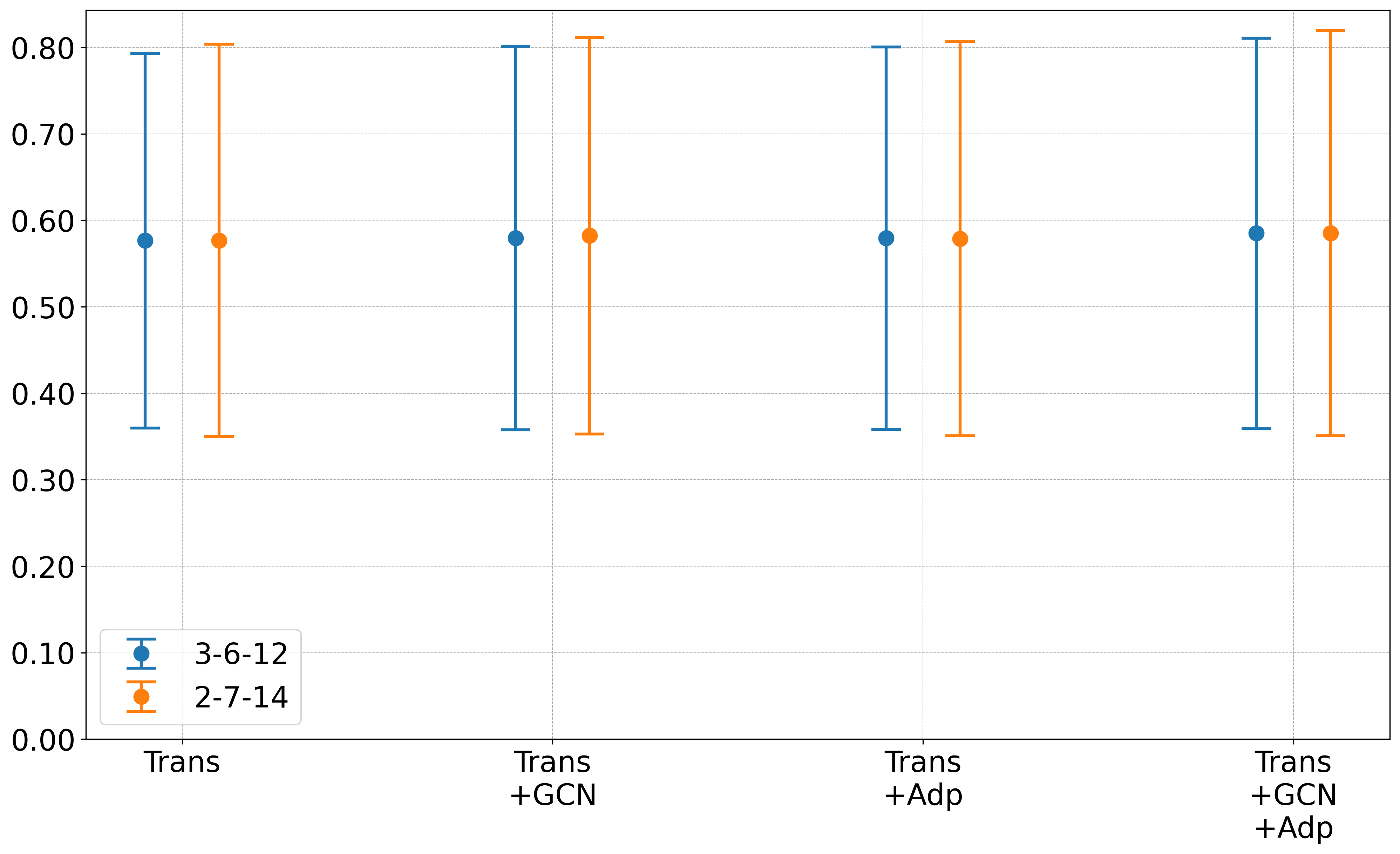}
         \caption{RMSE scores}
         \label{fig:weekly_rmse}
     \end{subfigure}
     \caption{\textbf{MAE and RMSE scores for incidence forecasting with and without weekly information.}}
     \label{fig:weekly_trend}
\end{figure}

\section*{Main experiments: Analysis on map connectivity via mobility indicator}
\subsection*{Analysis on average non-zero weights}
Besides connectivity, the average value of non-zero weights is also worth investigating. We find that static maps for EU $(0.454)$ and US $(0.491)$ contain weights of higher values than in the dynamic maps for EU $(0.151)$ and US $(0.358)$, with the difference being more significant in the EU maps. The difference between dynamic maps from Trans+Adp and Trans+GCN+Adp is less conspicuous for EU $(0.121, 0.180)$ and US $(0.404, 0.312)$, showing that with or without static maps, dynamic maps can still retrieve mobility information. The EU's highest average weight from Trans+Adp is $0.127$ for output length $6$, and from Trans+GCN+Adp is $0.215$ for output length $24$. For the US, the highest average weight from Trans+Adp is $0.500$ for output length $24$, and from Trans+GCN+Adp is $0.467$ for output length $6$. From the fold perspective, fold 1 for output length $24$ generates the highest average weight $0.163$ for EU Trans+Adp. Fold 5 for output length $3$ offers the highest average weight $0.272$ for EU Trans+GCN+Adp. For the US Trans+Adp, fold 3 with output length $6$ generates the highest average weight of $0.929$. For the US Trans+GCN+Adp, fold 1 with output length $6$ generates the highest average weight of $0.822$. 

\subsection*{Analysis on least and most connected sub-regions}
The connectivities for subregions for both the EU and the US are also explored. We define the subregion with the minimum and maximum connectivity in its matrix as one vote. Since subregions can have the same amount of edges, both minimum and maximum votes can refer to several subregions. Maps reduced to identity matrices are excluded. For the EU Trans+Adp, Denmark, with 13 votes, is recognized as the least connected node, and Austria and Belgium, with 8 votes, are the most connected nodes. For EU Trans+GCN+Adp, Germany, with 13 votes, is the least connected node and Austria and Bulgaria, with 11 votes are the most connected. For US Trans+Adp, Arizona, California, and Colorado, with 15 votes are voted the least connected, and Indiana, with 4 votes, is voted the most connected. For US Trans+GCN+Adp, Arkansas, with 17 votes, is voted the least connected, and California, with 8 votes, is voted as the most connected. We notice that the dynamic maps from Trans+GCN+Adp and Trans+Adp do not reach agreement on the subregion connectivity, offering limited insight on the subregional level. 

\subsection*{Relationship between lockdown and mobility indicator}
The relationship between lockdown orders and mobility indicator is summarized in Supplementary Table ~\ref{tab:lockdown_mobility_eu} and Table ~\ref{tab:lockdown_mobility_us}.

\begin{table}[h!]
    \centering
    \caption{\textbf{EU Mobility Indicator for Pre-Lockdown, Lockdown, and Post-Lockdown period. }Bold indicates a decrease in indicators.}
    \label{tab:lockdown_mobility_eu}
    \begin{tabular}{llp{15mm}p{15mm}p{15mm}p{15mm}p{15mm}}
    \toprule
    Subregion &
    Lockdown Period &
    Lockdown Indicator &
    Pre-Lockdown Indicator Average &
    Pre-Lockdown Indicator First &
    Post-Lockdown Indicator Average &
    Post-Lockdown Indicator Last \\
    \midrule
    Austria    & Mar.16,2020-Apr.30,2020  & 0.454 & \bf{0.492} & \bf{0.508} & \bf{0.505} & \bf{0.495} \\
    Austria    & Nov.17,2020-Dec.6,2020   & 0.385 & 0.363 & 0.637 & 0.269 & 0.731 \\
    Austria    & Dec.26,2020-Feb.7,2021   & 0.355 & 0.230 & 0.770 & \bf{0.394} & \bf{0.606} \\
    Austria    & Nov.15,2021-Dec.26,2021  & 0.357 & 0.253 & 0.747 & 0.203 & 0.797 \\
    Austria    & Jan.1,2022-Jan.31,2022   & 0.257 & \bf{0.321} & \bf{0.679} & \bf{0.260} & \bf{0.740} \\
    Belgium    & Mar.18,2020-May.9,2020   & 0.460 & \bf{0.482} & \bf{0.518} & \bf{0.489} & \bf{0.511} \\
    Cyprus     & Mar.24,2020-May.3,2020   & 0.462 & \bf{0.471} & \bf{0.529} & \bf{0.495} & \bf{0.505} \\
    Cyprus     & Jan.10,2021-May.9,2021   & 0.404 & 0.343 & 0.657 & 0.378 & 0.622 \\
    Greece     & Mar.23, 2020-May.4,2020  & 0.445 & 0.447 & \bf{0.553} & \bf{0.467} & 0.533 \\
    Spain      & Mar.14,2020-Apr.23,2020  & 0.437 & \bf{0.458} & \bf{0.542} & \bf{0.484} & \bf{0.516} \\
    France     & Mar.17, 2020-May.11,2020 & 0.440 & \bf{0.467} & \bf{0.533} & 0.439 & 0.561 \\
    France     & Oct.29,2020-Dec.14,2020  & 0.321 & \bf{0.329} & \bf{0.671} & 0.287 & \bf{0.713} \\
    Croatia    & Mar.23, 2020-May.10,2020 & 0.450 & 0.447 & \bf{0.553} & 0.441 & 0.559 \\
    Hungary    & Mar.27, 2020-May.18,2020 & 0.453 & 0.437 & \bf{0.563} & 0.425 & 0.575 \\
    Ireland    & Mar.27, 2020-May.17,2020 & 0.453 & 0.437 & \bf{0.563} & 0.428 & 0.572 \\
    Ireland    & Oct.21,2020-Nov.30,2020  & 0.319 & \bf{0.327} & 0.673 & 0.301 & 0.699 \\
    Ireland    & Dec.31,2020-Apr.11,2021  & 0.281 & 0.264 & 0.736 & \bf{0.364} & \bf{0.636} \\
    Italy      & Mar.10,2020-May.4,2020   & 0.436 & \bf{0.476} & \bf{0.524} & \bf{0.467} & \bf{0.533} \\
    Lithuania  & Dec.16,2020-Mar.31,2021  & 0.274 & \bf{0.342} & \bf{0.658} & \bf{0.395} & \bf{0.605} \\
    Luxembourg & Mar.18,2020-Apr.9,2020   & 0.411 & \bf{0.481} & \bf{0.519} & 0.411 & \bf{0.589} \\
    Poland     & Mar.24,2020-Apr.9,2020   & 0.428 & \bf{0.449} & \bf{0.551} & \bf{0.493} & \bf{0.507} \\
    Portugal   & Mar.19,2020-May.2,2020   & 0.430 & \bf{0.479} & \bf{0.521} & \bf{0.464} & \bf{0.536} \\
    Portugal   & Jan.15,2021-Apr.30,2021  & 0.282 & \bf{0.292} & 0.708 & \bf{0.349} & \bf{0.651} \\
    Czechia    & Mar.16,2020-Apr.24,2020  & 0.421 & \bf{0.483} & \bf{0.517} & \bf{0.483} & \bf{0.517} \\
    Czechia    & Oct.22,2020-Nov.3,2020   & 0.228 & \bf{0.346} & \bf{0.654} & \bf{0.417} & \bf{0.583} \\
    \bottomrule
    \end{tabular}
\end{table}

\begin{table}[h!]
    \centering
    \caption{\textbf{US Mobility Indicator for Pre-Lockdown, Lockdown, and Post-Lockdown period. } Bold indicates a decrease in indicators.}
    \label{tab:lockdown_mobility_us}
    \begin{tabular}{llp{15mm}p{15mm}p{15mm}p{15mm}p{15mm}}
    \toprule
    Subregion &
    Lockdown Period &
    Lockdown Indicator &
    Pre-Lockdown Indicator Average &
    Pre-Lockdown Indicator First &
    Post-Lockdown Indicator Average &
    Post-Lockdown Indicator Last \\
    \midrule
    Alabama        & Apr.4,2020-Apr.29,2020   & 0.372 & 0.364          & 0.255 & 0.364          & \textbf{0.429} \\
    Arizona        & Mar.31,2020-May.10,2020  & 0.386 & 0.318          & 0.255 & \textbf{0.446} & \textbf{0.456} \\
    California     & Mar.19,2020-May.11,2020  & 0.380 & 0.255          & 0.255 & \textbf{0.441} & \textbf{0.454} \\
    Colorado       & Mar.26,2020-Apr.26,2020  & 0.390 & 0.296          & 0.255 & 0.387          & \textbf{0.458} \\
    Delaware       & Mar.24,2020-May.21,2020  & 0.393 & 0.266          & 0.255 & \textbf{0.441} & \textbf{0.477} \\
    DC &
      Apr.1,2020-May.28,2020 &
      0.397 &
      0.321 &
      0.255 &
      \textbf{0.481} &
      \textbf{0.507} \\
    Florida        & Apr.3,2020-May.3,2020    & 0.370 & 0.357          & 0.255 & \textbf{0.407} & \textbf{0.410} \\
    Georgia        & Apr.3,2020-Apr.26,2020   & 0.374 & 0.357          & 0.255 & \textbf{0.387} & \textbf{0.458} \\
    Idaho          & Mar.25,2020-Apr.30,2020  & 0.389 & 0.294          & 0.255 & 0.361          & \textbf{0.419} \\
    Illinois       & Mar.21,2020-May.28,2020  & 0.382 & 0.255          & 0.255 & \textbf{0.481} & \textbf{0.507} \\
    Indiana        & Mar.24,2020-May.3,2020   & 0.377 & 0.266          & 0.255 & \textbf{0.407} & \textbf{0.410} \\
    Kansas         & Mar.30,2020-May.3,2020   & 0.377 & 0.305          & 0.255 & \textbf{0.407} & \textbf{0.410} \\
    Louisiana      & Mar.23,2020-May.14,2020  & 0.391 & 0.263          & 0.255 & \textbf{0.423} & \textbf{0.458} \\
    Maine          & Apr.2,2020-May.30,2020   & 0.399 & 0.352          & 0.255 & \textbf{0.479} & \textbf{0.495} \\
    Maryland       & Mar.30,2020-May.12,2020  & 0.366 & 0.305          & 0.255 & \textbf{0.438} & \textbf{0.456} \\
    Michigan       & Mar.24,2020-May.20,2020  & 0.393 & 0.266          & 0.255 & \textbf{0.446} & \textbf{0.480} \\
    Minnesota      & Mar.27,2020-May.16,2020  & 0.389 & 0.298          & 0.255 & \textbf{0.437} & \textbf{0.500} \\
    Mississippi    & Apr.3,2020-Apr.26,2020   & 0.374 & 0.357          & 0.255 & \textbf{0.387} & \textbf{0.458} \\
    Missouri       & Apr.6,2020-May.3,2020    & 0.365 & \textbf{0.372} & 0.255 & \textbf{0.407} & \textbf{0.410} \\
    Montana        & Mar.28,2020-Apr.25,2020  & 0.394 & 0.299          & 0.255 & 0.383          & \textbf{0.458} \\
    Nevada         & Mar.31,2020-May.8,2020   & 0.374 & 0.318          & 0.255 & 0.374          & \textbf{0.393} \\
    New Hampshire  & Mar.27,2020-May.17,2020  & 0.389 & 0.298          & 0.255 & \textbf{0.442} & \textbf{0.491} \\
    New Jersey     & Mar.21,2020-Jun.8,2020   & 0.392 & 0.255          & 0.255 & \textbf{0.493} & \textbf{0.502} \\
    North Carolina & Mar.30,2020-May.21,2020  & 0.373 & 0.305          & 0.255 & \textbf{0.441} & \textbf{0.477} \\
    Ohio           & Mar.23,2020-May.14,2020  & 0.391 & 0.263          & 0.255 & \textbf{0.423} & \textbf{0.458} \\
    Oregon         & Mar.23,2020-May.14,2020  & 0.391 & 0.263          & 0.255 & \textbf{0.423} & \textbf{0.458} \\
    Pennsylvania   & Apr.1,2020-May.7,2020    & 0.368 & 0.321          & 0.255 & \textbf{0.370} & 0.368          \\
    Rhode Island   & Mar.28,2020-May.8,2020   & 0.377 & 0.299          & 0.255 & 0.374          & \textbf{0.393} \\
    South Carolina &
      Apr.7,2020-May.3,2020 &
      0.357 &
      \textbf{0.373} &
      0.255 &
      \textbf{0.407} &
      \textbf{0.410} \\
    Tennessee      & Apr.2,2020-Apr.26,2020   & 0.380 & 0.352          & 0.255 & \textbf{0.387} & \textbf{0.458} \\
    Vermont        & Mar.24,2020-May.14,2020  & 0.387 & 0.266          & 0.255 & \textbf{0.423} & \textbf{0.458} \\
    Virginia       & Mar.30,2020-May.14,2020  & 0.366 & 0.305          & 0.255 & \textbf{0.423} & \textbf{0.458} \\
    Washington     & Mar.23,2020-May.7,2020   & 0.382 & 0.263          & 0.255 & 0.370          & 0.368          \\
    West Virginia  & Mar.24,2020-May.3,2020   & 0.377 & 0.266          & 0.255 & \textbf{0.407} & \textbf{0.410} \\
    Wisconsin      & Mar.25,2020-May.12,2020  & 0.394 & 0.294          & 0.255 & \textbf{0.438} & \textbf{0.456} \\
    \bottomrule
    \end{tabular}
\end{table}

\subsection*{Relationship between lockdown  and forecasting performances}
\begin{figure}[ht!]
     \centering
     \begin{subfigure}[b]{0.45\textwidth}
         \centering
         \includegraphics[width=\textwidth]{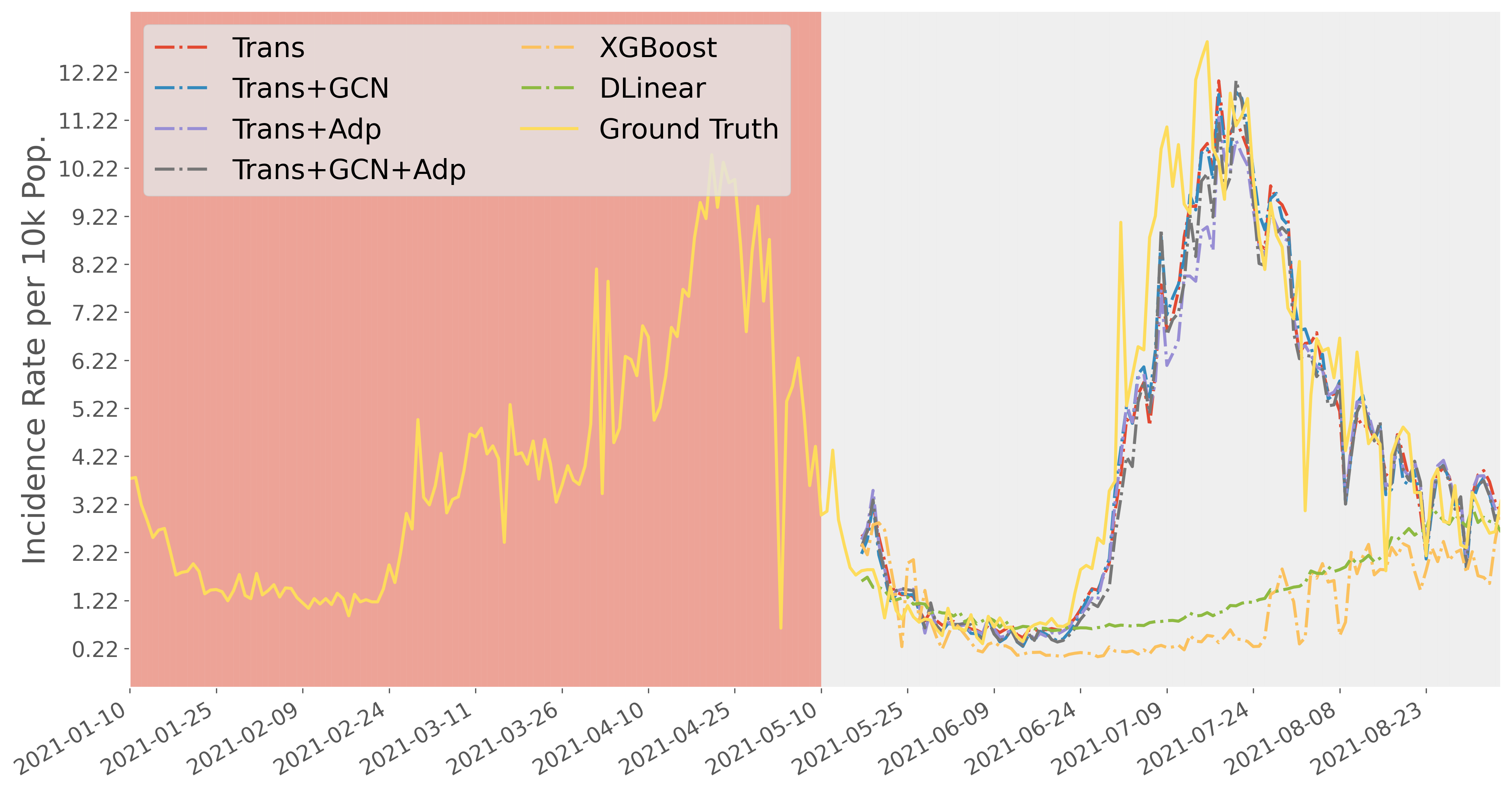}
         \caption{Forecasting of EU Cyprus from May 17, 2021 to Sep 05, 2021 (Fold 2)}
         \label{fig:LD_EU}
     \end{subfigure}
     \hfill
     \begin{subfigure}[b]{0.45\textwidth}
         \centering
         \includegraphics[width=\textwidth]{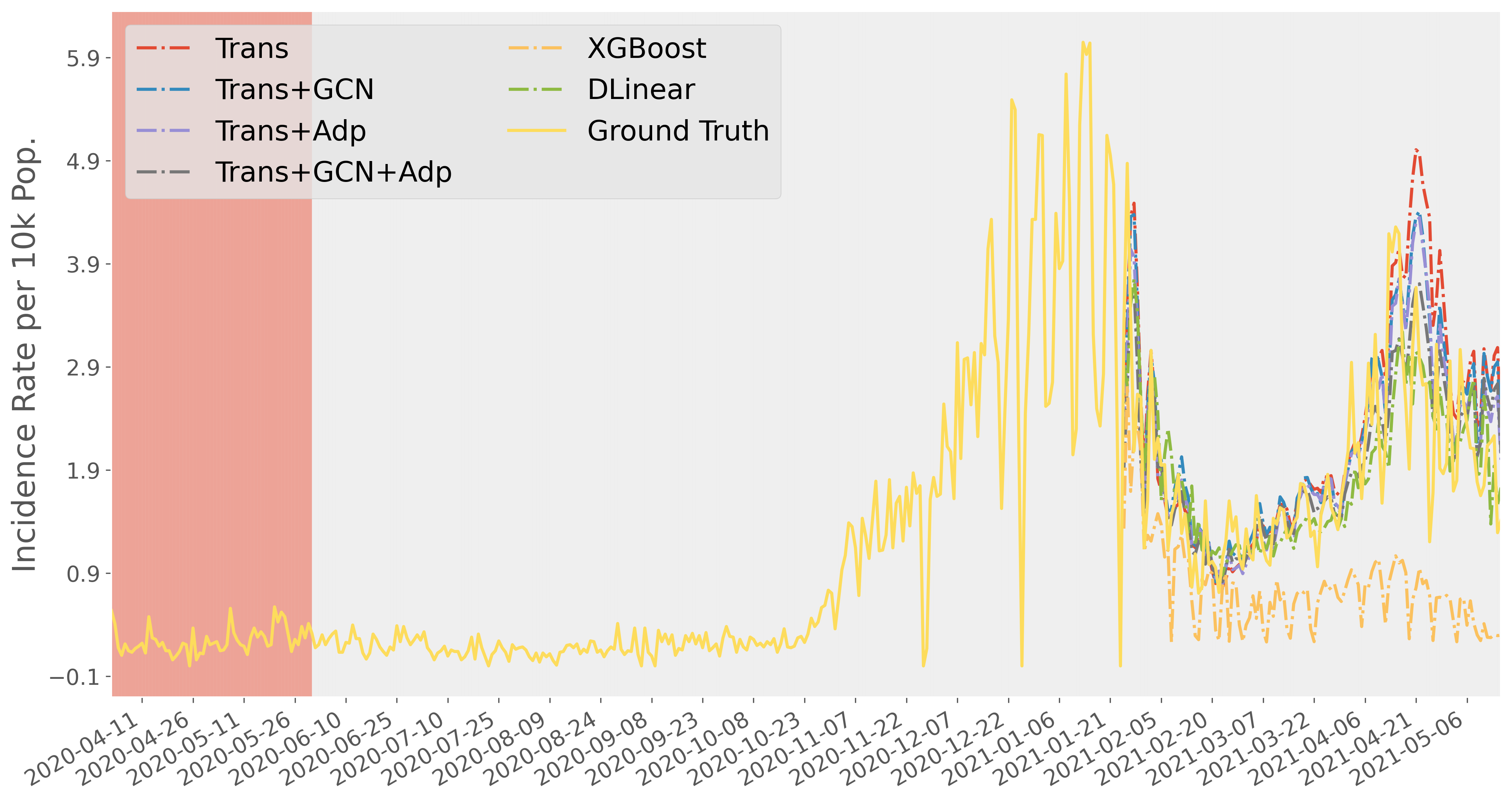}
         \caption{Forecasting of US Maine from Jan 25, 2021 to May 16, 2021 (Fold 1)}
         \label{fig:LD_US}
     \end{subfigure}
        \caption{\textbf{Impact of Lockdown orders on incidence forecasting.} Results of XGBoost, DLinear, and our models with $12$ output timesteps are reported, showcasing their ability to address the influence of NPIs on incidence. The salmon-color-coded region indicates the enforcement of lockdown measures.}
        \label{fig:LockDown}
\end{figure}

Regarding the duration of lockdown periods, our models reliably predict the subsequent rebound in cases, as shown in Fig. \ref{fig:LD_EU} for the EU and Fig. \ref{fig:LD_US} for the US. In the case of the US, most lockdown orders were issued at the beginning of the pandemic, and only the test set from fold $1$ can capture their lingering effect. In contrast, the EU sites issued multiple lockdown orders spanning the years 2020 and 2021, enabling visualization of forecasting across different folds. Fig. \ref{fig:LockDown} shows that models that recognize spatial correlations, explicitly or implicitly, respond better to post-lockdown rebounds, suggesting that spatial connections are informative for post-lockdown incidence forecasting. We observe that the transformer backbone provides reliable forecasting, and the addition of GCN modules does not seem to improve the prediction, which is consistent with our findings.

\section*{Descriptive figures}
\begin{figure}[h]
\centering
\includegraphics[width=0.8\linewidth]{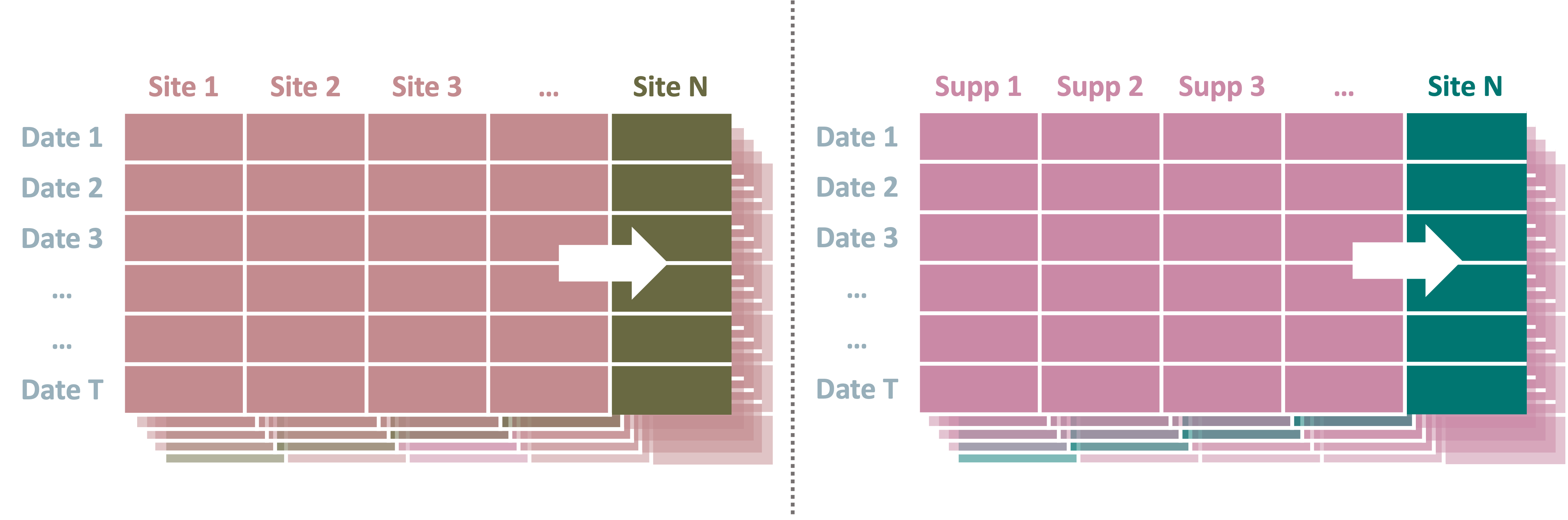}
\caption{\textbf{The location-wise and supplementary-wise prediction for shallow machine learning algorithms.} The left table shows location-wise prediction input data with daily incidence values, while the right table represents supplementary-wise prediction input data with daily supplementary variable values. The last column denotes the daily incidence at the selected location.}
\label{fig:extended_shallow_input}
\end{figure}

In Supplementary Fig.\ref{fig:extended_triad}, Fig. \ref{fig:extended_fold_split}
and Fig. \ref{fig:extended_shallow_input}, we have created visualizations of the correlation between incidence, mortality, and hospitalization, the progressive split strategy to create 5-fold data, and the dataset organization strategy for location-wise and supplementary-wise prediction for shallow learning algorithms.

\begin{figure}[h]
\centering
\includegraphics[width=0.7\linewidth]{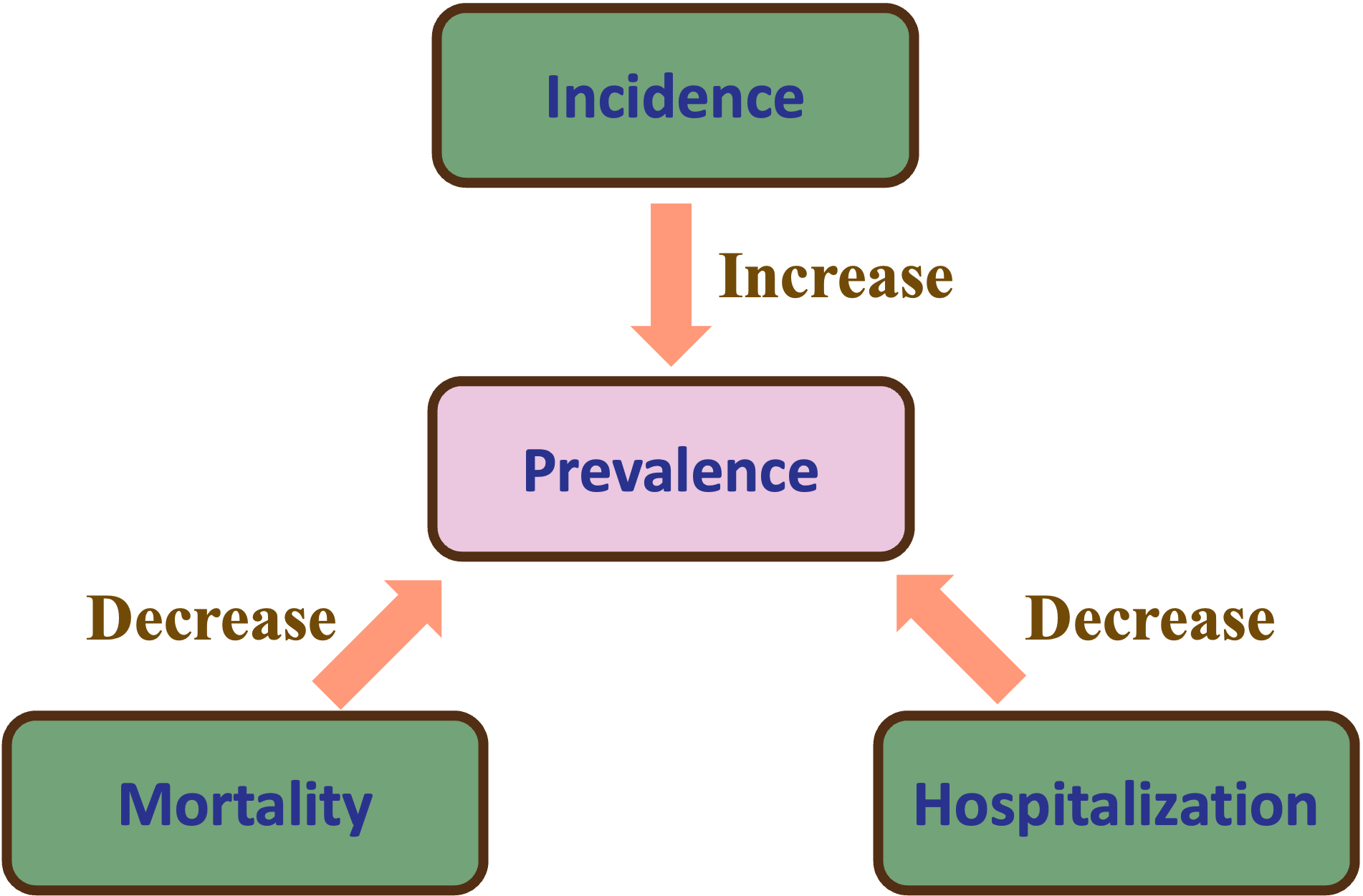}
\caption{\textbf{The relationship between incidence, mortality, and hospitalization rate (a proxy for recovered).} The growth in incidence increases future prevalence and the increments in mortality and hospitalization rate decrease future prevalence.}
\label{fig:extended_triad}
\end{figure}

\begin{figure}[h]
\centering
\includegraphics[width=0.6\linewidth]{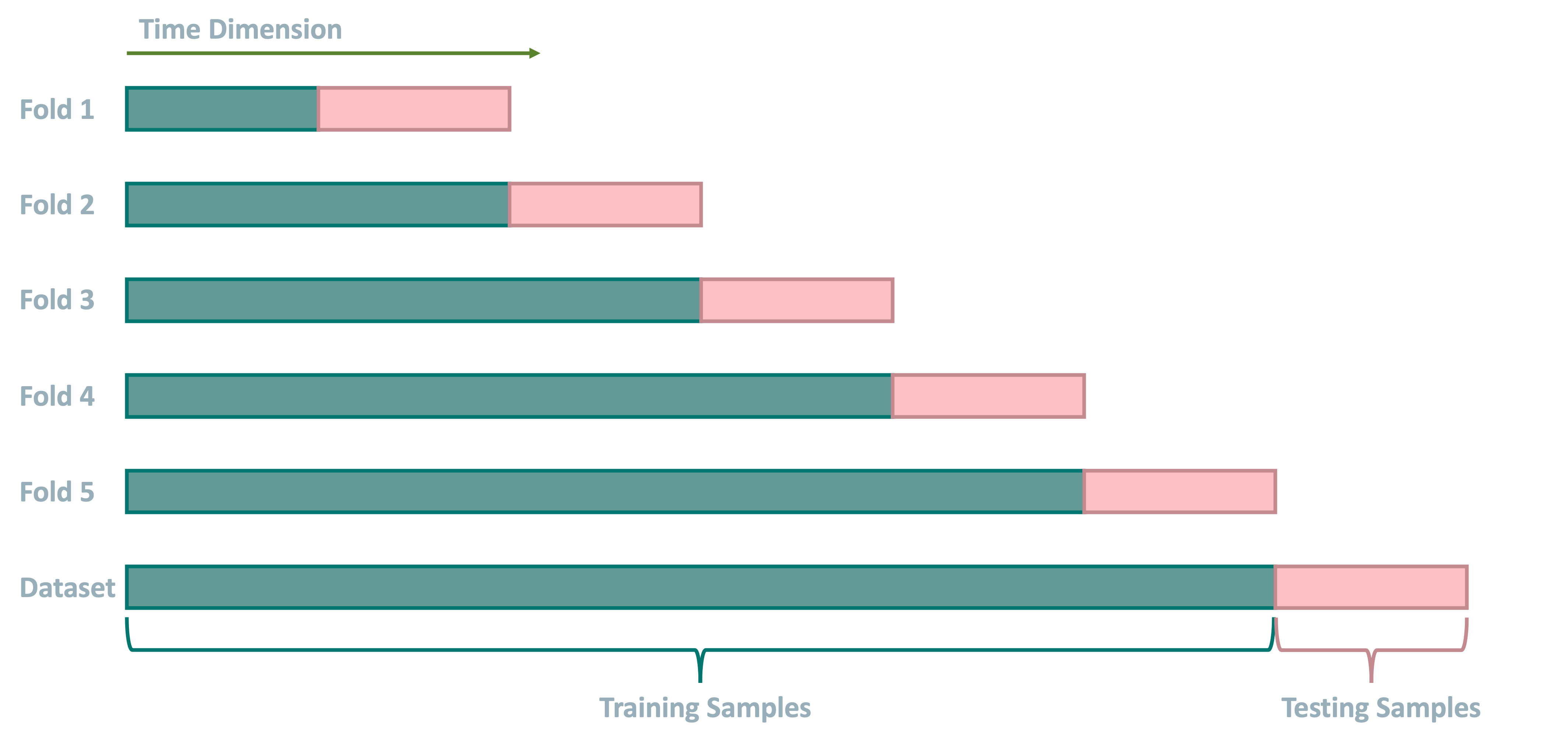}
\caption{\textbf{The $5$-fold data split strategy.} Training instances are split chronologically. The best-fitted model in each fold is tested on the holdout samples.}
\label{fig:extended_fold_split}
\end{figure}

\section*{Imputation, Notations and hyperparameters}
In Supplementary Table \ref{tab:extented_imputation} and Table \ref{tab: extented_mobility_missing}, we have summarized the missingness of data in incidence, mortality, and hospitalization rates as well as mobility data. We observe an exceptional amount of missing in the EU hospitalization rate. 

\begin{table}[]
\centering
\caption{Imputation count in days for EU and US}
\label{tab:extented_imputation}
\begin{tabular}{lllll|llll}
\toprule
Region & Sub-Region& I  & M  & H   & Sub-Region     & I  & M  & H \\
\midrule
\multirow{14}{*}{EU} & Austria & 2  & 2  & 70  & Italy & 1  & 1  & 36  \\
   & Belgium       & 0  & 1  & 56  & Latvia         & 0  & 1  & 247 \\
   & Bulgaria      & 0  & 0  & 82  & Lithuania      & 1  & 7  & 531 \\
   & Croatia       & 0  & 0  & 182 & Luxembourg     & 1  & 2  & 50  \\
   & Cyprus        & 0  & 0  & 244 & Malta          & 1  & 1  & 351 \\
   & Czechia       & 5  & 13 & 49  & Netherlands    & 1  & 6  & 109 \\
   & Denmark       & 1  & 2  & 74  & Norway         & 0  & 2  & 386 \\
   & Estonia       & 1  & 2  & 26  & Poland         & 0  & 0  & 312 \\
   & Finland       & 0  & 0  & 240 & Portugal       & 1  & 0  & 267 \\
   & France        & 12 & 7  & 58  & Romania        & 0  & 0  & 411 \\
   & Germany       & 2  & 3  & 911 & Slovakia       & 2  & 1  & 51  \\
   & Greece        & 0  & 0  & 934 & Slovenia       & 0  & 1  & 80  \\
   & Hungary       & 0  & 0  & 262 & Spain          & 5  & 8  & 196 \\
   & Ireland       & 4  & 7  & 55  & Sweden         & 0  & 7  & 297 \\
\midrule
\multirow{24}{*}{US} & Alabama  & 2  & 6  & 0   & Nebraska & 10 & 32 & 0   \\
   & Arizona       & 0  & 21 & 0   & Nevada         & 4  & 5  & 0   \\
   & Arkansas      & 2  & 5  & 0   & New Hampshire  & 1  & 1  & 0   \\
   & California    & 2  & 13 & 0   & New Jersey     & 4  & 10 & 0   \\
   & Colorado      & 0  & 3  & 0   & New Mexico     & 0  & 1  & 0   \\
   & Connecticut   & 2  & 22 & 0   & New York       & 0  & 7  & 0   \\
   & Delaware      & 6  & 2  & 0   & North Carolina & 0  & 2  & 0   \\
   & District of Columbia & 1 & 4 & 0 & North Dakota & 0 & 5  & 0   \\
   & Florida       & 5  & 3  & 0   & Ohio           & 0  & 0  & 0   \\
   & Georgia       & 1  & 5  & 0   & Oklahoma       & 2  & 3  & 0   \\
   & Idaho         & 2  & 11 & 0   & Oregon         & 0  & 1  & 0   \\
   & Illinois      & 3  & 7  & 0   & Pennsylvania   & 1  & 2  & 0   \\
   & Indiana       & 0  & 1  & 0   & Rhode Island   & 5  & 1  & 0   \\
   & Iowa          & 2  & 1  & 0   & South Carolina & 5  & 7  & 0   \\
   & Kansas        & 1  & 18 & 0   & South Dakota   & 2  & 1  & 0   \\
   & Kentucky      & 1  & 1  & 0   & Tennessee      & 0  & 11 & 0   \\
   & Louisiana     & 1  & 0  & 0   & Texas          & 1  & 5  & 0   \\
   & Maine         & 10 & 4  & 0   & Utah           & 0  & 1  & 0   \\
   & Maryland      & 2  & 4  & 0   & Vermont        & 2  & 1  & 0   \\
   & Massachusetts & 1  & 2  & 0   & Virginia       & 3  & 8  & 0   \\
   & Michigan      & 1  & 4  & 0   & Washington     & 9  & 14 & 0   \\
   & Minnesota     & 0  & 1  & 0   & West Virginia  & 1  & 7  & 0   \\
   & Mississippi   & 0  & 0  & 0   & Wisconsin      & 0  & 10 & 0   \\
   & Missouri      & 3  & 19 & 0   & Wyoming        & 4  & 4  & 0   \\
   & Montana       & 2  & 7  & 0   &                &    &    &     \\
\bottomrule
\end{tabular}
\end{table}

\begin{table}[]
\centering
\caption{Ratio of missing mobility data for EU and US}
\label{tab: extented_mobility_missing}
\begin{tabular}{lllllll}
\toprule
Region & retail\_and\_recreation & grocery\_and\_pharmacy & parks & transit\_stations & residential & workplaces \\
\midrule
EU     & 0.272                   & 0.291                  & 0.509 & 0.339             & 0.288       & 0.063      \\
US     & 0.289                   & 0.372                  & 0.666 & 0.547             & 0.351       & 0.021     \\
\bottomrule
\end{tabular}
\end{table}

In Table \ref{tab:extended_notations}, we have outlined the notations used in the Method section in the main article. The shape and description of the notation can be found accordingly. Test set sizes are documented in Table \ref{tab:extended_sample_size_timesteps}, and hyper-parameters for the shallow-machine-learning algorithms are reported in Table \ref{tab:extended_shallow_param}.

\begin{table}[h]
\centering
\caption{Notations}
\begin{tabular}{@{}c|c@{}} 
 \toprule
 Notation & Description \\ 
 \hline
 $T, N, F$ & window size, number of sites/nodes, forecasting horizon  \\ 
 $D$, $D_k$, $D_v$ & feature dimensions \\
 $\tau$, $\rho$ & pre-set threshold and sparsity ratio \\
 $\mathbf{X} \in \mathbb{R}^{N \times T}$ & training input for $N$ locations \\
 $\mathbf{x}_i \in \mathbb{R}^{1 \times T} $ & training input for location $i$ \\
 $\mathbf{Y} \in \mathbb{R}^{N \times F}$ & training target for $N$ locations \\
 $\mathbf{A}_g \in \mathbb{R}^{N \times N}$ & geographical adjacency matrix \\
 $\mathbf{A}_s^{(l)} \in \mathbb{R}^{N \times N} $ & generated adjacency matrix for block $l$ \\ 
 $ \mathbf{s} \in \mathbb{R}^{1 \times N} $ & $i$-th row vector of nodes attention matrix \\
 $\mathbf{M}_n \in \mathbb{R}^{N \times N} $ & node attention score \\
 $\mathbf{Z}^{(l)} \in \mathbb{R}^{N \times T \times D}$ &  input embedding for block $l$ \\
 $\mathbf{Z}_t^{(l)} \in \mathbb{R}^{N \times T \times D}$ &  temporal embedding block $l$ \\
 $\mathbf{h}^{(l)}$ & graph embedding for block $l$ \\
 \bottomrule
\end{tabular}
\label{tab:extended_notations}
\end{table}

\begin{table}[ht]
\centering
\caption{Sample Size and Timesteps for Test Sets}
\begin{tabular}{@{}ccccl@{} } 
 \toprule
 Window Size &  Forecasting Horizon & Sample Size & Timesteps & Dates(dd/mm/yy)\\
 \cmidrule[0.4pt](r{0.125em}){1-1}
 \cmidrule[0.4pt](r{0.125em}){2-2}
 \cmidrule[0.4pt](r{0.125em}){3-3}
 \cmidrule[0.4pt](r{0.125em}){4-4}
 \cmidrule[0.4pt](r{0.125em}){5-5}
 
 \multirow{3}{*}{$12$} & $3$ & $102$ & $104$ & $08/16/22 - 11/27/22$ \\
 & $6$ & $102$ & $107$ & $08/13/22 - 11/27/22$ \\
 & $12$ & $101$ & $112$ & $08/08/22 - 11/27/22$ \\
 & $24$ & $100$ & $122$ & $07/28/22 - 11/27/22$ \\
 & $36$ & $99$ & $134$ & $07/17/22 - 11/27/22$ \\
 
 \cmidrule[0.4pt](r{0.125em}){1-1}
 \cmidrule[0.4pt](r{0.125em}){2-2}
 \cmidrule[0.4pt](r{0.125em}){3-3}
 \cmidrule[0.4pt](r{0.125em}){4-4}
 \cmidrule[0.4pt](r{0.125em}){5-5}
 
 \multirow{3}{*}{$14$} & $2$ & $102$ & $103$ & $08/17/22 - 11/27/22$ \\
 & $7$ & $102$ & $108$ & $08/12/22 - 11/27/22$\\
 & $14$ & $101$ & $114$ & $08/06/22 - 11/27/22$\\
 
 \bottomrule
\end{tabular}
\label{tab:extended_sample_size_timesteps}
\end{table}

\begin{table}[h!]
\centering
\caption{Shallow Machine Learning Parameters}
\begin{tabular}{@{}lccc@{}} 
 \toprule
 Parameter Name & Values Tested $1$ & Values Tested $2$ & Values Tested $3$\\ 
 \cmidrule[0.4pt](r{0.125em}){1-1}
 \cmidrule[0.4pt](r{0.125em}){2-2}
 \cmidrule[0.4pt](r{0.125em}){3-3}
 \cmidrule[0.4pt](r{0.125em}){4-4}
 max\_depth & $5$ & $30$ & $55$ \\
 n\_estimators & $100$ & $500$ & $900$ \\
 learning\_rate & $0.1$ & $0.01$ & $0.05$ \\
 max\_features\footnotemark[1] & $1.0$ & sqrt & log2 \\
 \bottomrule
\end{tabular}
\label{tab:extended_shallow_param}
\footnotetext[1]{This parameter is only used in the RandomForest Model.}
\end{table}